\documentclass{article}

\usepackage{microtype}
\usepackage{graphicx}
\usepackage{subfigure}
\usepackage{booktabs} % for professional tables
\usepackage{lipsum}
\usepackage{enumitem}
\usepackage{import}

\usepackage{hyperref}

\usepackage[accepted]{icml2023}

\usepackage{amsmath}
\usepackage{amssymb}
\usepackage{mathtools}
\usepackage{amsthm}

\usepackage[capitalize,noabbrev]{cleveref}

\theoremstyle{plain}
\newtheorem{theorem}{Theorem}[section]
\newtheorem{proposition}[theorem]{Proposition}
\newtheorem{lemma}[theorem]{Lemma}
\newtheorem{corollary}[theorem]{Corollary}
\theoremstyle{definition}
\newtheorem{definition}[theorem]{Definition}

\theoremstyle{remark}

\usepackage[textsize=tiny]{todonotes}

\usepackage{amsmath}
\usepackage{amssymb}
\usepackage{amsthm}
\usepackage{mathtools}
\usepackage{xcolor}
\usepackage{hyperref}
\usepackage{graphicx}
\usepackage{bm}

\usepackage{tikz}

\newlength{\indentationFormule} 
\newlength{\indentationTotaleFormule}
\newlength{\indentationCommentaire}
\newlength{\indentationDerivation}
\newlength{\largeurLangle}
\newlength{\largeurBoiteCommentaire}
{\setlength{\indentationDerivation}{#1}%
\setlength{\indentationTotaleFormule}{\indentationFormule}
\addtolength{\indentationTotaleFormule}{#1}
\setlength{\largeurBoiteCommentaire}{\linewidth}
\addtolength{\largeurBoiteCommentaire}{-\indentationFormule}
\addtolength{\largeurBoiteCommentaire}{-\indentationCommentaire}
\addtolength{\largeurBoiteCommentaire}{-\largeurLangle}
\addtolength{\largeurBoiteCommentaire}{-\indentationDerivation}
\begin{list}{}{\setlength{\leftmargin}{\indentationTotaleFormule}}
\setlength{\baselineskip}{2.3\baselineskip}
\item$}%
{\hbox{}$\end{list}}  % \hbox{} evite des erfull lors d'usages
\newcommand{\somme}[3]{\sum_{#1}^{#2}{#3}}

\newcommand{\integral}[4]{\int_{#1}^{#2} #3 \,#4}

\newcommand{\real}{\mathbb{R}}

\newcommand{\theset}[1]{\{ #1 \}}
\newcommand{\lipschitz}[1]{\text{\rm Lip}_{#1} }

\newcommand{\loss}{\ell}
\newcommand{\xfancy}{\mathcal{X}}

\newcommand{\zfancy}{\mathcal{Z}}
\newcommand{\dfancy}{\mathcal{D}}
\newcommand{\hfancy}{\mathcal{H}}

\newcommand{\ffancy}{\mathcal{F}}
\newcommand{\gfancy}{\mathcal{G}}

\newcommand{\wfancy}{\mathcal{W}}

\newcommand{\rfancy}{\mathcal{R}}
\newcommand{\sfancy}{\mathcal{S}}

\newcommand{\zeromatrix}{\mathbf{0}}
\newcommand{\idmatrix}{\mathbf{I}}

\newcommand{\zvector}{\mathbf{z}}
\newcommand{\yvector}{\mathbf{y}}
\newcommand{\xvector}{\mathbf{x}}

\newcommand{\equdef}{\stackrel{\text{def}}{=}}
\newcommand{\sampled}{\sim}

\newcommand{\absolu}[1]{\left \lvert #1 \right \rvert}

\newcommand{\expon}[1]{e^{#1}}
\newcommand{\exponbig}[1]{\textnormal{exp}\left[ #1    \right]}

\newcommand{\supremum}[2]{\sup_{#1} #2}

\newcommand{\pushf}[2]{{#1} \sharp #2}

\newcommand{\prob}[2]{\underset{#1}{\mathbb{P}}\left[#2\right]}
\DeclareMathOperator*{\Esp}{\mathbb{E}}
\newcommand{\expect}[2]{\Esp_{#1}\left[#2\right]}
\newcommand{\expectseul}[2]{\Esp_{#1} #2 }
\newcommand{\expecttext}[2]{\Esp_{#1}\,[#2]}

\newcommand{\normal}[1]{\mathcal{N}(#1)}
\newcommand{\mprob}[1]{\mathcal{M}_+^1(#1)}
\newcommand{\kl}[2]{\mathrm{KL}(#1 \, || \, #2)}

\newcommand{\abscont}{\ll}

\newcommand{\listen}[1]{#1_1, \dots, #1_n}

\newcommand{\bleu}[1]{\textcolor{blue}{#1}}
\newcommand{\details}[1]{\bleu{#1}}

\newcommand{\andspace}{\quad \text{ and } \quad}
\newcommand{\unsur}[1]{\frac{1}{#1}}
\newcommand{\omet}[1]{\details{$\bullet \bullet \bullet \bullet $}}
\newcommand{\finpreuve}{\hfill $\square$}
\newcommand{\guillemets}[1]{``#1''}

\newcommand{\tv}{d_{TV}}

\newcommand{\dist}[1]{P^{#1}}

\newcommand{\unfunc}[1]{\mathbf{1}_{[#1]}}

\newcommand{\diam}[1]{\text{diam}(#1)}
\newcommand{\dimz}{{d_\zfancy}}

\newcommand{\wprime}{W_\ffancy}
\newcommand{\tvprime}{\dfancy_\ffancy}
\newcommand{\ipm}[1]{d_{#1}}
\newcommand{\otimesn}{^{\otimes n}}
\newcommand{\latentdist}{P_\zfancy}

\newcommand{\emprisk}[1]{\hat{\rfancy}_S(#1)}
\newcommand{\truerisk}[1]{\rfancy(#1)}

\newcommand{\newrisk}[1]{\wfancy_\ffancy\left( #1  \right)}
\newcommand{\newtv}[1]{\dfancy_\ffancy\left( #1  \right)}
\newcommand{\newwprime}[1]{d_\ffancy(#1)}

\begin{document}

\twocolumn[
%\icmltitle{PAC-Bayesian Guarantees for Adversarial Generative Models}
\icmltitle{PAC-Bayesian Generalization Bounds for Adversarial Generative Models}




\begin{icmlauthorlist}
\icmlauthor{Sokhna Diarra Mbacke}{xxx}
\icmlauthor{Florence Clerc}{yyy}
\icmlauthor{Pascal Germain}{xxx}
\end{icmlauthorlist}

\icmlaffiliation{xxx}{Université Laval}
\icmlaffiliation{yyy}{McGill University}
%\icmlaffiliation{sch}{School of ZZZ, Institute of WWW, Location, Country}

\icmlcorrespondingauthor{Sokhna Diarra Mbacke}{sokhna-diarra.mbacke.1@ulaval.ca}
\icmlcorrespondingauthor{Florence Clerc}{florence.clerc@mail.mcgill.ca}
\icmlcorrespondingauthor{Pascal Germain}{pascal.germain@ift.ulaval.ca}

\icmlkeywords{PAC-Bayes, Generative Models, Statistical Learning Theory, Wasserstein GANs}

\vskip 0.3in
]



\printAffiliationsAndNotice{}  % leave blank if no need to mention equal contribution

\begin{abstract}
We extend PAC-Bayesian theory to generative models and develop generalization bounds for models based on the Wasserstein distance and the total variation distance. Our first result on the Wasserstein distance assumes the instance space is bounded, while our second result takes advantage of dimensionality reduction.
Our results naturally apply to Wasserstein GANs and Energy-Based GANs, and our bounds provide new training objectives for these two. 
Although our work is mainly theoretical, we perform numerical experiments showing non-vacuous generalization bounds for Wasserstein GANs on synthetic datasets. 

\end{abstract}

\section{Introduction} \label{sec-intro}
Deep Generative models have become a central research area in machine learning. Two of the most popular families of deep generative models are Variational Autoencoders (VAEs) \citep{autoencoding, rezende-vae} and Generative Adversarial Networks (GANs) \citep{gan-goodfellow}. GANs are known for producing impressive results in image generation \citep{large-scale, style-gan}, generating fake images indistinguishable from real ones. They also have been applied to video \citep{video-gan}, text \cite{text-gan-survey} and protein generation \citep{protein-gan}.

\paragraph{Motivation.} In this work, we study the generalization properties of GANs using PAC-Bayesian theory. Considering the prevalence of GANs in machine learning, the question of generalization is important for numerous reasons. First, quantitatively measuring the discrepancy between the generator's distribution and the true distribution is a difficult problem. Indeed, there are known issues with the current evaluation metrics \citep{note-eval-gen-models, pros-cons}, and it can be quite challenging to detect when the generator only produces slight variations of the training samples. Moreover, having generalization bounds not only contributes to the theoretical understanding of GANs themselves, but also to the understanding of the structure of real-life datasets, if those can be provably approximated by GAN-generated data. In addition, given that GANs are used for data-augmentation in fields such as medical image classification \citep[see e.g.][]{frid-adar}, theoretical guarantees can substantiate the soundness of such applications.

\subsection{Notations and Preliminaries.} 
The set of $K$-Lipschitz functions defined on a space $\xfancy$ is denoted $\lipschitz{K}$ and the set of probability measures on $\xfancy$ is denoted $\mprob{\xfancy}$. Integral Probability Metrics {\citep[IPM, see][]{ipm-def}} are a class of pseudometrics\footnote{For the sake of readability, we will call also call pseudometrics distances in this work.} defined on the space of probability measures. Given $P, Q \in \mprob{\xfancy}$ and a space $\ffancy$ of real-valued functions defined on $\xfancy$, the IPM induced by $\ffancy$ is defined as
\begin{equation}\label{eq-ipm}
\ipm{\ffancy}(P, Q) = \sup_{f\in \ffancy}\absolu{ \integral{}{ }{f}{dP} - \integral{}{}{f}{dQ}  }.
\end{equation}
Examples of IPMs include the total variation distance $\tv$, corresponding to the case \mbox{$\ffancy = \theset{f\,{:}\, \xfancy \rightarrow \real: -1 \leq f \leq 1}$} and the Wasserstein distance $W_1$, corresponding to $\ffancy {\,=\,} \lipschitz{1}$.

\paragraph{Generative Adversarial Networks.} GANs have two main components: the generator $g \in \gfancy$ and the critic $f \in \ffancy$, where both $\gfancy$ and $\ffancy$ are parameterized by neural networks. Given a $n$-sized training set $S = \theset{\listen{\xvector}}$ iid sampled from an unknown distribution $\dist{*}$ on a space $\xfancy$, the generator is trained to produce samples that \guillemets{look like} they came from $\dist{*}$ and the critic is trained to tell apart the real samples from the fake ones. The original GAN of \citet{gan-goodfellow} has been shown to minimize the Jensen-Shannon divergence (JS) between the true distribution $\dist{*}$ and the generator's distribution, denoted $\dist{g}$. 

The original GAN suffers from many problems such as training instability and mode collapse \citep{improved-gan-training}. Upon providing some theoretical explanations for these issues, \citet{wgan} introduce the Wasserstein GAN (WGAN), which replaces JS by the Wasserstein-1 distance \citep{villani} between $\dist{*}$ and $\dist{g}$. Thanks to the Kantorovich-Rubinstein duality, the minimization of $W_1(\dist{*}, \dist{g})$ is equivalent to the following objective:
\begin{equation}\label{eq-wgan_}
\min_{g} \max_{f \in \lipschitz{1}} \left\{ \expect{\xvector \sampled \dist{*}}{f(\xvector)} - \expect{\hat{\xvector} \sampled \dist{g}}{f(\hat{\xvector})}     \right\}.
\end{equation}
%where $\lipschitz{1}$ denotes the sets of $1$-Lipschitz functions defined on $\xfancy$. 
In practice, however, $\lipschitz{1}$ is replaced by a family $\ffancy$ of neural networks referred to as the \emph{critic family}. This leads to the following objective
\begin{equation}\label{eq-wgan-practice_}
\min_{g} d_\ffancy(\dist{*}, \dist{g}),
\end{equation}
where $d_\ffancy$ is sometimes referred to as the neural divergence or neural IPM \citep{gen-equilibrium, insights}, since $\ffancy$ is a family of neural networks.

Another variant of GANs is the Energy-based GAN \citep{ebgan}, which views the critic as an energy function and uses a margin loss. More precisely, given a positive number $m$ called the margin, EBGAN's critic and generator minimize respectively
\[
\min_{f} \left\{ \expectseul{\xvector \sampled \dist{*}}{f(\xvector)} + \expectseul{\hat{\xvector} \sampled \dist{g}}{\max \left(0, m - f(\hat{\xvector}) \right)} \right\},
\]
and
\[
\min_{g} \left\{ \expectseul{\hat{\xvector} \sampled \dist{g}}{f(\hat{\xvector})} - \expectseul{\xvector\sampled \dist{*}}{f(\xvector)} \right\}.
\]
Note that the critic is constrained to be non-negative. 
\citet{wgan} showed that under an optimal critic, the EBGAN's generator minimizes (a constant scaling of) the total variation distance $\tv(\dist{*}, \dist{g})$.

\paragraph{Generalization.}Since the true distribution $\dist{*}$ is unknown and the model has only access to its empirical counterpart $\dist{*}_n$, the question of generalization naturally arises: How to certify that the learned distribution $\dist{g}$ is \guillemets{close} to the true one  $\dist{*}$? The goal of this work is to study the generalization properties of GANs using PAC-Bayesian theory. More precisely, we prove non-vacuous PAC-Bayesian generalization bounds for generative models based on the Wasserstein distance and the total variation distance. Since we use the IPM formulation of these metrics, our results are naturally applicable to WGANs and EBGANs.

\subsection{Related Works}There is a large body of works dedicated to the understanding of the generalization properties of GANs \citep{gen-equilibrium, disc-gen-tradeoff, liang, singh, uppal, without-domination, insights}. Given a family of generators $\gfancy$, a family of critics~$\ffancy$, and a discrepancy measure $\dfancy$, the usual goal is to upper bound the quantity $\dfancy(\dist{*}, \dist{{\hat{g}}})$, where $\hat{g}$ is an optimal solution to the empirical problem $\min_{g\in \gfancy}\dfancy(\dist{*}_n, \dist{g})$. From a statistical perspective, the most common approach is to quantify the rate of convergence of $r(\hat{g})\coloneqq \dfancy(\dist{*}, \dist{{\hat{g}}}) - \inf_{g\in \gfancy}\dfancy(\dist{*}, \dist{g})$, as the size of the training set $n$ goes to infinity. Assuming that the target distribution $\dist{*}$ has a smooth density, \citet{singh, liang} and \citet{uppal} provide rates of convergence dependent on the ambient dimension of the instance space $\xfancy$ and the complexity of the critic family $\ffancy$. Noting that the density assumption on $\dist{*}$ might be unrealistic in practice, \citet{without-domination} prove rates of convergence assuming $\dist{*}$ is a smooth transformation of the uniform distribution on a low-dimensional manifold. This allows them to derive rates depending on the intrinsic dimension of the data, as opposed to its extrinsic dimension. Under simplicity assumptions on the critic family, \citet{disc-gen-tradeoff} provide upper bounds for $r(\hat{g})$, when $\dfancy$ is the negative critic loss $d_\ffancy$. They first prove general bounds using the Rademacher complexity of $\ffancy$, then bound this complexity in the case when $\ffancy$ is a family of neural networks with certain constraints. More recently, \citet{insights} developed upper bounds for $r(\hat{g})$, but assuming $\dfancy$ is the Wasserstein-1 distance $W_1$. They argue that since the use of $d_\ffancy$ in practice is purely motivated by optimization considerations, $W_1$ is a better way of assessing the generalization properties of WGANs.

One major distinction between this work and the ones cited above, is that our definition of the generalization error does not explicitly involve the modeling error $\inf_{g\in \gfancy}\dfancy(\dist{*}, \dist{g})$. Instead, we define the generalization error as the discrepancy between the empirical loss and the expected population loss, allowing us to derive bounds that can be turned into an optimization objective to be minimized by a learning algorithm. Our approach to generalization is closer to the one taken by \citet{gen-equilibrium}, who study the generalization properties of GANs by defining the generalization error, for any generator $g$, as $\absolu{\dfancy(\dist{*}, \dist{g}) - \dfancy(\dist{*}_n, \dist{g}_n)}$, 
where $\dfancy(\dist{*}_n, \dist{g}_n)$ is the discrepancy between the empirical training and generated distributions.
%for any generator $g$. 
They show that models minimizing $W_1$ do not generalize (in the sense that the generalization error cannot be made arbitrarily small, given a polynomial number of samples), while models minimizing $d_\ffancy$ do, under certain conditions on $\ffancy$. A distinction between our approach and the one taken by \citet{gen-equilibrium} is that we define the empirical risk as the expectation $\expectseul{}{\dfancy(\dist{*}_n, \dist{g}_n)}$ with respect to the fake distribution $\dist{g}_n$, since in practice, the samples defining $\dist{g}_n$ are drawn anew at each iteration. Moreover, we study distributions $\rho \in \mprob{\gfancy}$ over the set of generators, as well as individual generators $g \in \gfancy$.

There are other differences between our approach and the ones above. First, our bounds do not depend on the complexity or smoothness of the critic family $\ffancy$. In other words, our generalization bounds apply systematically to any critic family $\ffancy$, with no distinctions between the cases where $\ffancy$ is a \guillemets{small} subset of $\lipschitz{1}$ and where $\ffancy = \lipschitz{1}$. The intuitive explanation is that the complexity of the critic family is naturally \guillemets{embedded} in the empirical and population risks defined in the PAC-Bayesian framework. Second, because of the generality of the PAC-Bayesian theory, we make no assumptions on the structure of the critic family, and some of our bounds do not even make assumptions on the hypothesis space $\gfancy$. The fact that these results can be directly applied to neural networks is a consequence of the generality of PAC-Bayes bounds. Moreover, our bounds provide novel training objectives, giving rise to models that use the training data to not only learn the distribution $\dist{*}$, but also obtain a risk certificate valid on previously unseen data.

Aside from the study of the generalization properties of GANs, our work relates to the recent work of \citet{shedding}, who develop PAC-Bayes bounds for \guillemets{adaptative} sliced Wasserstein distances. The sliced-Wasserstein distance (SW) \citep{sliced-wasserstein} is an optimization-focused alternative to the Wasserstein distance. Given distributions $P$ and $Q$ on a high-dimensional space, SW computes $W_1(P_1, Q_1)$ instead of $W_1(P, Q)$, where $P_1$ and $Q_1$ are projections of $P$ and $Q$ on a 1-dimensional space. 
Note that the bounds developed by \citet{shedding} apply to the SW distance, whereas our bounds are developed for the Wasserstein distance between distributions on a high dimensional space. In addition, the bounds of \citet{shedding} focus on the discriminative setting, that is, the models they study optimize to find the projections with the highest discriminative power. Then, they argue that these bounds can be applied to the study of generative models based on the distributional sliced-Wasserstein \cite{dsw}. In contrast, our results are specifically tailored to the generative modeling setting and provide upper bounds on the difference between the empirical risk of a critic and its population risk.

Finally, we mention a recent article \citep{pb-vae} which uses PAC-Bayes to obtain generalization bounds on the \emph{reconstruction loss} of VAEs. In short, \citet{pb-vae} clip the reconstruction loss in order to utilize McAllester's bound \citep{mc2003a}, which applies to $[0, 1]$-bounded loss functions. Moreover, they omit the KL-loss, meaning they do not analyze a VAE per se, but simply a stochastic reconstruction machine. Hence, theirs is not a PAC-Bayesian analysis of a generative model, but of a reconstruction model.

\subsection{Our Contributions}
The primary objective of this work is to extend PAC-Bayesian theory to adversarial generative models. We develop novel PAC-Bayesian generalization bounds for generative models based on the Wasserstein distance and the total variation distance. 
First, assuming the instance space is bounded, we prove generalizations bounds for Wasserstein models dependent on the diameter of the instance space. Then, we show that one can obtain bounds dependent on the intrinsic dimension, assuming that the distributions are smooth transformations of a distribution on a low-dimensional space. Finally, we exhibit generalization bounds for models based on the total variation distance. 
To the best of our knowledge, ours are the first PAC-Bayes bounds developed for the generalization properties of generative models. 
Our results naturally apply to Wasserstein GANs and Energy-Based GANs. Moreover, our bounds provide new training objectives for WGANs and EBGANs, leading to models with statistical guarantees. It is noteworthy that we make no density assumptions on the true and generated distributions. Although our main motivation is theoretical, we perform numerical experiments showing non-vacuous generalization bounds for WGANs on synthetic datasets. We also report the results of preliminary experiments on the MNIST dataset.

\section{PAC-Bayesian Theory}

PAC-Bayesian theory \citep[introduced by][]{some_pb_thms} applies Probably Approximately Correct (PAC) inequalities to \emph{pseudo-Bayesian} learning algorithms---whose output could be framed as a \emph{posterior} probability distribution over a class of candidate models---
%\footnote{By \emph{pseudo-Bayesian} learning algorithm, we mean that its output could be framed as a \emph{posterior} probability distribution over a class of candidate models.} 
in order to provide generalization bounds for machine learning models. Here, the term generalization bound refers to upper bounds on the discrepancy between a model's empirical loss and its population loss (\emph{i.e.}, the loss on the true data distribution). Optimizing these bounds lead to \emph{self-certified} learning algorithms, that produce models whose behavior on the population is statistically guaranteed to be close to their behavior on the observed samples.  PAC-Bayes has been applied to a wide variety of settings such as classification \citep{pac-bayes-linear, pb-classif}, linear regression \citep{meets, pb-lin-regression}, meta-learning \citep{pb-meta-learning}, variational inference for mixture models \citep{pb-vi-mixture} and online learning \citep{online-pb}. In recent years, PAC-Bayes has been used to obtain non-vacuous generalization bounds for neural networks \citep{diff-privacy, perez-ortiz}. See \citet{primer} and \citet{friendly} for recent surveys.

The wide variety of applications is due to the flexibility of the PAC-Bayesian framework. 
Indeed, the theory is very general, and requires few assumptions. We consider a training set $S =\theset{\listen{\xvector}}$, iid sampled from an unknown probability distribution $\dist{*}$ over an instance space~$\xfancy$.\footnote{A vast majority of the PAC-Bayes literature is devoted to the prediction setting where each training instance is a pair $(x, y)$ of some features $x$ and a label $y$. 
%a \emph{features-label} pair, that is $\xvector\coloneqq(x,y)$. 
We adopt slightly more general definitions that encompass unsupervised learning.}
Given a hypothesis class $\hfancy$ and a real-valued loss function $\loss: \hfancy \times \xfancy \rightarrow [0, \infty)$, the empirical and population risks of each hypothesis $h \in \hfancy$ are respectively defined as
$$\emprisk{h} = \unsur{n} \somme{i=1}{n}{\loss(h, \xvector_i)} \mbox{ \ and \  } \truerisk{h} = \expect{\xvector \sampled \dist{*}}{\loss(h, \xvector)}.$$ 
Instead of individual hypotheses $h\in \hfancy$, PAC-Bayes focuses on a \emph{posterior} probability distributions over hypotheses $\rho \in \mprob{\hfancy}$. These distributions can be seen as \emph{aggregate} hypotheses.
Similar to the risks for individual hypotheses, the empirical and true risks of an aggregate hypothesis \mbox{$\rho \in \mprob{\hfancy}$} are respectively defined as 
$$\emprisk{\rho} = \expecttext{h\sampled \rho}{\emprisk{h}}  \mbox{ \ and \  }\truerisk{\rho} = \expect{h\sampled \rho}{\truerisk{h}}.$$ 
The goal of PAC-Bayesian theory is to provide upper bounds on the discrepancy between $\truerisk{\rho}$ and $\emprisk{\rho}$ which hold with high probability over the random draw of the training set $S$. As an example, consider the following general PAC-Bayes bound originally developed by \citet{pac-bayes-linear} and further formalized by \citet{unleashed}.

\begin{theorem}\label{thm-gen-bound}
 Let $\pi \in \mprob{\hfancy}$ be a prior distribution independent of the data, $D : \real^+ \times \real^+ \rightarrow \real^+$ be a convex function, and $\delta \in (0, 1)$ be a real number. With probability at least $1-\delta$ over the random draw of $S \sampled {\dist{*}}\otimesn$, the following holds for any $\rho \in \mprob{\hfancy}$ such that $\rho \abscont \pi$ and $\pi \abscont \rho$:
\begin{equation}\label{eq-gen-pb-bound}
\begin{split}
D \left( \truerisk{\rho}, \emprisk{\rho}  \right) \leq &\ \kl{\rho}{\pi} + \log \unsur{\delta} \\ 
&{} + \log \expectseul{h \sampled \pi} \ {\expectseul{\mathrlap{\!\!\!\!S\sampled {\dist{*}}\otimesn}}{\expon{ D \left( \truerisk{h}, \emprisk{h}   \right)   }  }   }\,, 
\end{split}
\end{equation}
where $\kl{\rho}{\pi}$ is the Kullback-Leibler divergence between distributions $\rho$ and $\pi$.
%where $\expectseul{S}{}$ abbreviates $\expectseul{S \sampled {\dist{*}}\otimesn}{}$.
\end{theorem}
The left-hand side of Equation~\eqref{eq-gen-pb-bound} quantifies the discrepancy between the true risk $\truerisk{\rho}$ and its empirical counterpart $\emprisk{\rho}$ for a given training set $S$, while the complexity term of the right-hand side involves the expectation with respect to $S\sampled {\dist{*}}\otimesn$. As the data distribution ${\dist{*}}$ is unknown, the latter term needs to be upper-bounded in order to obtain a finite and numerically computable bound.

Theorem~\ref{thm-gen-bound} requires $\rho \abscont \pi$, which is classic in PAC-Bayes bounds and necessary for the KL-divergence to be defined. However, it also requires $\pi \abscont \rho$ which seems a bit more restrictive. As noted by \citet{unleashed}, one has to make sure that $\pi$ and $\rho$ have the same support, which is the case when they are from the same parametric family of distributions, such as Gaussian or Laplace. Although the KL-divergence appears in most PAC-Bayes bounds, some bounds have been developed with the Rényi divergence \citep{pb-renyi} and IPMs \citep{pb-ipm}.

Finally, note that Theorem \ref{thm-gen-bound} requires the prior distribution $\pi$ to be independent of the training set $S$. Even though this restriction makes it easier to bound the exponential moment \citep{beyond-usual}, it may also lead to large values of the KL term in practice, since the posterior is likely to be far from the prior. A common strategy is to use a portion of the training data to learn the prior, while making sure this portion is not used in the numerical computation of the bound \citep{perez-ortiz}.

Aside from bounds for aggregate hypotheses $\rho \in \mprob{\hfancy}$, PAC-Bayes bounds can be formulated for individual hypotheses $h\in \hfancy$ as well. Such bounds hold with high probability over the random draw of a single predictor $h$ sampled from the PAC-Bayesian posterior, and have appeared in, e.g., \citet{catoni-07}. In some cases, the derandomization step is quite straightforward, as a result of the structure of the hypotheses. For instance, \citet{pac-bayes-linear} utilize the linearity of the hypotheses to express a randomized linear classifier as a single deterministic linear classifier. In the general case, however, it can be quite challenging and costly to derandomize PAC-Bayesian bounds \citep{neyshabur, nagarajan,biggs-guedj}. Below, we present a result by \citet{beyond-usual}, who provide a general theorem for derandomizing PAC-Bayes bounds.

\begin{theorem}\label{thm-gen-derand}
With the definitions and assumptions of Theorem \ref{thm-gen-bound}, given a measurable function $f : \sfancy \times \hfancy \rightarrow \real$, the following holds with probability at least $1-\delta$ over the random draws of $S \sampled {\dist{*}}\otimesn$ and $h \sampled \rho$:
\begin{equation}\label{eq-gen-derand}
f(S, h) \leq \log \frac{d\rho}{d\pi}(h) + \log \unsur{\delta} + \log \expectseul{h \sampled \pi}{\expectseul{S\sampled {\dist{*}}\otimesn}{\expon{ f(S, h)  }  }   }.
\end{equation}
\end{theorem}
%Bounds in the shape of Equation~\eqref{eq-gen-derand} are known as \guillemets{derandomized} or \guillemets{desintegrated} bounds in the PAC-Bayes literature. 
Removing the expectation with respect to the hypothesis space, is very useful in applications to neural networks \citep{viallard}. Theorem~\ref{thm-gen-derand} uses the Radon-Nikodym derivative of $\rho$ with respect to $\pi$, which can lead to high variance when the bound is used as an optimization objective for neural networks. \citet{viallard} empirically highlighted this phenomenon, and formulated a generic disintegrated bound where the Radon-Nikodym derivative is replaced by the Renyi-divergence between $\rho$ and $\pi$.

\section{PAC-Bayesian Bounds for Generative Models}

This section presents our main results. We consider a metric space $(\xfancy, d)$, an unknown probability measure $\dist{*}$ on~$\xfancy$ and a training set $S = \theset{\listen{\xvector}}$ iid sampled from~$\dist{*}$. The empirical counterpart of $\dist{*}$ defined by $S$ is denoted~$\dist{*}_n$.
%meaning $\dist{*}_n = \unsur{n}\somme{i=1}{n}{\delta_{\xvector_i}}$.  
We also consider a hypothesis space $\gfancy$ such that each generator $g\in \gfancy$ induces a probability measure $\dist{g}$ on $\xfancy$, from which fake samples 
$S_g = \theset{\listen{\hat\xvector}}\sim {\dist{g}}\otimesn$
are generated. 
%Thus, $\dist{g}_n = \unsur{n}\somme{i=1}{n}{\delta_{\hat\xvector_i}}$.
Thus, 
$$\dist{*}_n = \unsur{n}\somme{i=1}{n}{\delta_{\xvector_i}}
\mbox{ \ and \ }
\dist{g}_n = \unsur{n}\somme{i=1}{n}{\delta_{\hat\xvector_i}},
$$
where $\delta_{\xvector_i}$ is the Dirac measure on sample $\xvector_i$.

\subsection{Bounds for Wasserstein generative models}

Let us consider a subset $\ffancy \subseteq \lipschitz{1}$ that is \emph{symmetric}, meaning $f\in \ffancy$ implies $-f \in \ffancy$. We emphasize that $\ffancy$ can be a small subset of $\lipschitz{1}$, or the whole set $\lipschitz{1}$. Given a generator $g\in \gfancy$, we define its empirical risk as
\begin{equation}\label{eq-newrisk}
\newrisk{\dist{*}_n, \dist{g}} = \expect{S_g}{\newwprime{\dist{*}_n, \dist{g}_n}},
\end{equation}
where the expectation is taken with respect to the iid sample~$S_g$ that induces $\dist{g}_n$ and $\newwprime{\dist{*}_n, \dist{g}_n}$ is the IPM induced by~$\ffancy$ (Equation~\ref{eq-ipm}).

The generalization error is defined as $$\expect{S\sampled {\dist{*}}\otimesn}{\newrisk{\dist{*}_n, \dist{g}}} - \newrisk{\dist{*}_n, \dist{g}},$$ namely the difference between the population and empirical risks. These definitions can be extended to aggregate generators by taking the expectation according to $\rho \in \mprob{\gfancy}$.
%The fact that this quantity is not guaranteed to be positive is not a problem, since the upper-bounds provided are positive. 
The following theorem provides  bounds on the generalization error of both \ref{thm-pb-bound-rand} aggregate and \ref{thm-pb-bound-derand} individual generators.

\begin{theorem} \label{thm-pb-bound}
Let $\ffancy \subseteq \lipschitz{1}$ be a symmetric set of real-valued functions on $\xfancy$, $\Delta \coloneqq \sup_{\xvector, \xvector' \in \xfancy} d(\xvector, \xvector') < \infty$ be the diameter of $\xfancy$,  $\dist{*} \in \mprob{\xfancy}$ be the true data-generating distribution and $S \in \xfancy^n$ a $n$-sized iid sample from $\dist{*}$. Consider a set of generators $\gfancy$ such that each $g \in \gfancy$ induces a distribution $\dist{g}$ on $\xfancy$, a prior distribution $\pi$ over $\gfancy$, and real numbers $\lambda > 0$ and $\delta \in (0, 1)$.

\begin{enumerate}[label=(\roman*)]

\item \label{thm-pb-bound-rand} For any probability measure $\rho$ over $\gfancy$ such that $\rho \abscont \pi$ and $\pi \abscont \rho$, the following holds with probability at least $1-\delta$ over the random draw of $S$:
\begin{align}\label{eq-bound}
&\hspace{-1cm}\expectseul{g \sampled \rho}{ 
\expect{S}{ \newrisk{\dist{*}_n, \dist{g}} }   }   - 
\expect{g\sampled \rho}{ \newrisk{\dist{*}_n, \dist{g} } } \nonumber\\ 
&\hspace{.3cm}\leq\unsur{\lambda }     \left[ \kl{\rho}{\pi}   +\log \unsur{\delta}   \right] + \frac{\lambda \Delta^2}{4n}.
\end{align}

\item \label{thm-pb-bound-derand} For any probability measure $\rho$ over $\gfancy$ such that $\rho \abscont \pi$ and $\pi \abscont \rho$, the following holds with probability at least $1-\delta$ over the random draw of $S$ and $g \sampled \rho$:
\begin{align}\label{eq-bound-derand}
&\hspace{-1cm}\expect{S }{ \newrisk{\dist{*}_n, \dist{g}}  }     -  \newrisk{\dist{*}_n, \dist{g}} \nonumber\\  
&\leq \unsur{\lambda }     \left[ \log\frac{d\rho}{d\pi}(g) +\log \unsur{\delta}   \right] + \frac{\lambda \Delta^2}{4n}.
\end{align}
\end{enumerate}

\end{theorem}

\emph{Proof Idea.} We provide a detailed outline of the proof here. The full details can be found in the supplementary material (Section \ref{sec-thm-pb-bound}).

The proof of \ref{thm-pb-bound-rand} relies on a technical lemma (Lemma \ref{lem-w1-bounddif}). It is possible to view $\ipm{\ffancy}(\dist{*}_n, \dist{g}_n)$ as a function $\xfancy^{2n} \to \mathbb{R}$ as $\dist{*}_n$ (resp. $\dist{g}_n$) is the uniform distribution on $n$ samples that were selected according to $\dist{*}$ (resp. $\dist{g}$). Lemma~\ref{lem-w1-bounddif} states that $\ipm{\ffancy}(\dist{*}_n, \dist{g}_n)$ has the bounded differences property with bounds $\Delta / n$, meaning that if we were to change only one sample, the new value of $\ipm{\ffancy}(\dist{*}_n, \dist{g}_n)$ would differ by at most $\Delta / n$ (see Definition~\ref{def-bounddif}). The proof (provided in the appendix) uses properties of the $\sup$ and the fact that $\ffancy \subset \lipschitz{1}$.

We then use a result used to prove McDiarmid's inequality (Lemma \ref{lem-mcdiarmid}, previously used by \citet{shedding} for their bounds on the sliced Wasserstein distance)
%we also provide a proof by \citet{mcdiarmid-note} in the appendix) 
and Fubini's theorem to obtain that
\[ \expectseul{S }{[ Y ]} \leq \exponbig{\frac{\lambda^2 \Delta^2}{4n}},
\]
where 
\[
Y \coloneqq\!\! \expectseul{g \sampled \pi}{
\expect{S_g}{
\exponbig{\lambda \left(  \expect{\!S, S_g\!}{\ipm{\ffancy}(\dist{g}_n, P^*_n)}   - \ipm{\ffancy}(\dist{g}_n, P^*_n)   \right)}     
}
}\!\!.
\]
Then, Markov's inequality combined with this result yields that with probability at least $1 - \delta$ over the random draw of the training set $S$, $$Y \leq \frac{1}{\delta} \exponbig{\frac{\lambda^2 \Delta^2}{4n}}.$$

The rest of the proof follows the main steps of the proof of Theorem~\ref{thm-gen-bound}, as presented by \citet{unleashed}. We use the Radon-Nikodym derivatives to change the expectation over $g \sampled \pi$ into an expectation over $g \sampled \rho$. Applying $\log$ (a monotone increasing function) to the inequality and then using Jensen's inequality for concave functions, with some further rewriting, yields \ref{thm-pb-bound-rand}.

In order to obtain \ref{thm-pb-bound-derand}, we study $\xi = \log \expectseul{S}{[Y]}$. Similarly to what happens in the proof of \ref{thm-pb-bound-rand}, we have that $\xi \leq \frac{\lambda^2 \Delta^2}{4n}$. However, using Jensen's inequality for convex functions, we can exchange the expectation over $S_g$ and $\exp$ in the definition of $Y$ to yield a new inequality. Combining it with previous result $\xi \leq \frac{\lambda^2 \Delta^2}{4n}$, we obtain that \[
\log \expectseul{S}{\expectseul{g \sampled \pi}{\expon{    \lambda \left(   \expectseul{S}{ \expect{S_g}{\ipm{\ffancy} (\dist{*}_n, \dist{g}_n) } }  - { \expectseul{S_g}{ \ipm{\ffancy} (\dist{*}_n, \dist{g}_n) }} \right)     }}} \leq \frac{\lambda ^2 \Delta^2}{4n} .
\]
We then use 
the general desintegrated bound by \citet{beyond-usual} stated in Theorem~\ref{thm-gen-derand}.
We take \[
f(S, g) = \lambda \left(   \expect{S, S_g}{\ipm{\ffancy} (\dist{*}_n, \dist{g}_n)  }  - \expect{S_g}{ \ipm{\ffancy} (\dist{*}_n, \dist{g}_n) } \right).\]
Previously obtained inequality enables us to bound
\[ \log \expectseul{S}{\expect{g \sampled \pi}{\expon{f(S, g)}}} \leq \frac{\lambda ^2 \Delta^2}{4n}, \]
which gives us the desired result and concludes the proof of~\ref{thm-pb-bound-derand}.
\finpreuve

Note that our desintegrated bound \eqref{eq-bound-derand} still has the expectation with respect to the fake sample $S_g$. Unlike the usual PAC-Bayesian bounds which are mostly applicable to supervised learning, the loss we are bounding requires not only some data from the unknown distribution, but also some data depending on the hypotheses. 

Theorem~\ref{thm-pb-bound} requires the samples $S_g$ from the generated distribution to have the same size $n$ as the training set. In practice, this is not a problem, since the user can easily sample from $\dist{g}$. One might wonder, however, if the bounds could be improved by increasing the number of fake samples. In our approach, the answer is no. Indeed, if the size of $S_g$ is $m \neq n$, then we obtain bounds with last term $\frac{\lambda \Delta^2}{4n}$ replaced by $ \frac{\lambda \Delta^2}{4 \min(m, n)}$.

Although Theorem~\ref{thm-pb-bound} provides upper bounds on the expected distance between empirical measures, it also implies upper bounds on the distance between the full distributions, as shown in the following corollary.
\begin{corollary}\label{cor-bound}
With the definitions and assumptions of Theorem~\ref{thm-pb-bound}, the following properties hold for any probability measure $\rho$ such that $\rho \abscont \pi$ and $\pi \abscont \rho$.
\begin{enumerate}[label=(\roman*)]

\item \label{cor-rand1} With probability at least $1-\delta$ over the random draw of $S$:
\begin{align*}
\hspace{-2mm}
\expectseul{g\sampled \rho}{\ipm{\ffancy}(\dist{*}, \dist{g}) } 
\leq&
\expect{g\sampled \rho}{ \newrisk{\dist{*}_n, \dist{g} } } \\  
&{} + \unsur{\lambda }     \left[ \kl{\rho}{\pi}   +\log \unsur{\delta}   \right] + \frac{\lambda \Delta^2}{4n}.
\end{align*}

\item \label{cor-derand1} With probability at least $1-\delta$ over the random draw of $S$ and $g \sampled \rho$:
\begin{align*}
\ipm{\ffancy}(\dist{*}, \dist{g})     \leq
& \ \newrisk{\dist{*}_n, \dist{g}} \\ 
& {}+\unsur{\lambda }     \left[ \log\frac{d\rho}{d\pi}(g) +\log \unsur{\delta}   \right] + \frac{\lambda \Delta^2}{4n}.
\end{align*}
\end{enumerate}
\end{corollary}
The proof of Corollary \ref{cor-bound} is in the supplementary material (Section \ref{sec-thm-pb-bound}). As a special case, when $\ffancy = \lipschitz{1}$, Corollary~\ref{cor-bound} provides upper bounds on the Wasserstein distance between the full distributions $\dist{*}$ and $\dist{g}$.

\paragraph{The manifold assumption.}
The bounds of Theorem \ref{thm-pb-bound} depend on the diameter of the instance space, which can be a handicap for real-world datasets such as image datasets. Indeed, the manifold hypothesis states that most high-dimensional real-world datasets lie in the vicinity of low-dimensional manifolds. There is a vast body of work dedicated to testing this assumption and estimating the intrinsic dimension of commonly used datasets \citep{survey-dimension-reduction, manifold-hypo, test-manifold-1, intrinsic-dim-images}. Moreover, latent variable generative models such as VAEs \cite{autoencoding}, GANs \cite{gan-goodfellow} and their variants exploit the manifold hypothesis by learning models which approximate distributions over high-dimensional spaces with transformations of low-dimensional latent distributions. This is also a main assumption of \citet{without-domination}, whose rates of convergence are dependent on the intrinsic dimension of the instance space. Taking a similar approach, we show that by assuming that the true distribution is a smooth transformation of a latent distribution over a low-dimensional hypercube, we can prove a PAC-Bayesian bound depending on the intrinsic dimension.

Before stating our next result, we recall the definition of a pushforward measure.
\begin{definition}[Pushforward Measure]
Given measurable spaces $\xfancy$ and $\zfancy$, a probability measure $\latentdist$ over $\zfancy$, and a measurable function $g : \zfancy \rightarrow \xfancy$, the pushforward measure defined by $g$ and $\latentdist$ is the probability distribution $\pushf{g}{\latentdist}$ on~$\xfancy$ defined as
\[
\pushf{g}{\latentdist} (A) = \latentdist(g^{-1}(A)),
\]
for any measurable set $A \subseteq \xfancy$. In more practical terms, sampling $\xvector$ from $\pushf{g}{\latentdist}$ means sampling a latent vector \mbox{$\zvector \sampled \latentdist$} first, then setting  $\xvector = g(\zvector)$. For example, a GAN's generator defines a pushforward distribution.
\end{definition}

\begin{theorem}\label{thm-pb-bound-lipschitz}
Let $\dist{*} \in \mprob{\xfancy}$ be the true data-generating distribution and $S \in \xfancy^n$ a $n$-sized iid sample from $\dist{*}$. We consider a set of generators $\gfancy$ such that each $g \in \gfancy$ induces a distribution $\dist{g}$ on $\xfancy$, a prior distribution $\pi$ over $\gfancy$, and real numbers $\lambda > 0$ and $\delta \in (0, 1)$. We also consider a latent space $\zfancy = [0, 1]^\dimz$, a latent distribution $\latentdist$ on $\zfancy$, and a true generator $g^*: \zfancy \rightarrow \xfancy$ such that $\dist{*} = \pushf{g^*}{\latentdist}$ and each $g \in \gfancy$ is a function $g: \zfancy \rightarrow \xfancy$ with $\dist{g} = \pushf{g}{\latentdist}$. Finally, we assume $\gfancy \cup \theset{g^*} \subseteq \lipschitz{K}$ for some positive real number $K$.

\begin{enumerate}[label=(\roman*)]

\item \label{thm-pb-lip-rand} For any probability measure $\rho$ over $\gfancy$ such that $\rho \abscont \pi$ and $\pi \abscont \rho$, the following holds with probability at least $1-\delta$ over the random draw of $S$:
\begin{equation}\label{eq-bound-lip-rand}
\begin{split}
\hspace{-4mm}
\expectseul{g \sampled \rho}{ 
\expect{S}{ \newrisk{\dist{*}_n, \dist{g}} }   }   - 
\expect{g\sampled \rho}{ \newrisk{\dist{*}_n, \dist{g} } } \\  
 \leq \unsur{\lambda }     \left[ \kl{\rho}{\pi}   +\log \unsur{\delta}   \right]  +  \frac{\lambda K^2 \dimz}{4n}  .
\end{split}
\end{equation}

\item \label{thm-pb-lip-derand} For any probability measure $\rho$ over $\gfancy$ such that $\rho \abscont \pi$ and $\pi \abscont \rho$, the following holds with probability at least $1-\delta$ over the random draw of $S$ and $g \sampled \rho$:
\begin{equation}\label{eq-bound-lip-derand}
\begin{split}
&\expect{S }{ \newrisk{\dist{*}_n, \dist{g}}  }      -  \newrisk{\dist{*}_n, \dist{g}} \\  
 &\qquad \leq \unsur{\lambda }     \left[ \log\frac{d\rho}{d\pi}(g) +\log \unsur{\delta}   \right] +   \frac{\lambda K^2 \dimz}{4n}  .
\end{split}
\end{equation}
\end{enumerate}
\end{theorem}

The proof can be found in Section \ref{sec-thm-pb-bound-lipschitz} in the appendix. The proof is very similar to that of \ref{thm-pb-bound-rand} of Theorem \ref{thm-pb-bound} but the technical lemma we rely on differs: instead of bounding small perturbations of $\ipm{\ffancy}(\dist{*}_n, \dist{g}_n)$ using the diameter $\Delta$, we bound those by $\frac{\lambda K^2 \dimz}{n}$ (see Lemma \ref{lem-w1-bounddif-lipschitz}).

As noted by \citet{without-domination}, the Lipschitz assumption on the true generator $g^*$ may be realistic in practice. Indeed, the generator learned by a GAN is a Lipschitz function of its input \citep{gan-lipschitz} and GAN-generated data has been shown to be a good substitute for real-life data in many applications \citep{frid-adar, low-shot,  data-aug-gan-ct, predicting-generalization-gan}. 

A result similar to Corollary~\ref{cor-bound} can be proven for Theorem~\ref{thm-pb-bound-lipschitz} (see Corollary~\ref{cor-bound-lipschitz}).

\subsection{Bounds for Total-Variation generative models}

In this section, we prove PAC-Bayesian generalization bounds for models based on the total variation distance. One such model is the EBGAN \cite{ebgan}. Indeed, \citet{wgan} show that given an optimal critic, the EBGAN's generator minimizes a constant scaling of the total variation distance between the real and fake distributions.

Let us assume $\ffancy$ is a symmetric set of functions $f {\,:\,} \xfancy {\rightarrow} [-1, 1]$ and denote $$\tvprime(\dist{*}_n, \dist{g}) = \expect{S_g}{\newwprime{\dist{*}_n, \dist{g}_n}}\,.$$
When $\ffancy$ is the set of all $[-1, 1]$-valued functions defined on $\xfancy$, then $\tvprime(\dist{*}_n, \dist{g})$ is the expected total variation distance between the real and fake empirical distributions.

\begin{theorem}\label{thm-pb-bound-tv}

Let $(\xfancy, d)$ be a metric space, $\dist{*} \in \mprob{\xfancy}$ be the true data-generating distribution and $S \in \xfancy^n$ a $n$-sized iid sample from $\dist{*}$. Consider a set of generators $\gfancy$ such that each $g \in \gfancy$ induces a distribution $\dist{g}$ on $\xfancy$, a prior distribution $\pi$ over $\gfancy$ and real numbers $\lambda > 0$ and $\delta \in (0, 1)$.

\begin{enumerate}[label=(\roman*)]

\item \label{thm-pb-bound-tv-rand} For any probability measure $\rho$ over $\gfancy$ such that $\rho \abscont \pi$ and $\pi \abscont \rho$, the following holds with probability at least $1-\delta$ over the random draw of $S$:
\begin{equation}\label{eq-bound-tv}
\begin{split}
\expectseul{g \sampled \rho}{ 
\expect{S}{ \newtv{\dist{*}_n, \dist{g}} }   }   - 
\expect{g\sampled \rho}{ \newtv{\dist{*}_n, \dist{g} } }\\  
 \leq  \unsur{\lambda }     \left[ \kl{\rho}{\pi}   +\log \unsur{\delta}   \right]  + \frac{4 \lambda}{n}.
\end{split}
\end{equation}

\item \label{thm-pb-bound-tv-derand} For any probability measure $\rho$ over $\gfancy$ such that $\rho \abscont \pi$ and $\pi \abscont \rho$, the following holds with probability at least $1-\delta$ over the random draw of $S$ and $g \sampled \rho$:

\begin{equation}\label{eq-bound-tv-derand}
\begin{split}
&\hspace{-6mm}\expect{S }{ \newtv{\dist{*}_n, \dist{g}}  }      -  \newtv{\dist{*}_n, \dist{g}}  \\  
&\quad\leq \unsur{\lambda }     \left[ \log\frac{d\rho}{d\pi}(g) +\log \unsur{\delta}   \right] + \frac{4 \lambda}{n}.
\end{split}
\end{equation}
\end{enumerate}

\end{theorem}

The proof of Theorem~\ref{thm-pb-bound-tv} is in the appendix (Section \ref{sec-thm-pb-bound-tv}).  The proof is very similar to that of \ref{thm-pb-bound-rand} of Theorem \ref{thm-pb-bound} but the technical lemma we rely on differs: instead of bounding small perturbations of $\ipm{\ffancy}(\dist{*}_n, \dist{g}_n)$ using the diameter $\Delta$, we bound those by $\frac{2}{n}$ (see Lemma \ref{lem-w1-bounddif-tv}).
A result similar to Corollary~\ref{cor-bound} for bounding the distance between the full distributions is also given by Corollary~\ref{cor-bound-tv}.

Note that unlike the bounds for the Wasserstein distance, the bounds for the total variation distance do not involve the size of the latent or instance space. This is not surprising, since $\tv$ can be seen as a special case of $W_1$ when the underlying metric on $\xfancy$ is $d = \unfunc{x \neq y}$. Results by \citet{wgan} show that the topology induced by the total variation distance is as strong as the one induced by the Jensen Shannon divergence, implying that EBGANs may suffer from some of the issues of the original GAN. Therefore, we focus our experiments on WGANs.

\subsection{Rate of convergence} 
The rate of convergence of the bounds proposed in this work depends on the choice of the hyperparameter $\lambda$. Choosing $\lambda =n$ leads to a fast rate of $n^{-1}$, but the bounds do not converge to $0$. The optimal rate for a convergence to $0$ is $n^{-1/2}$ and is obtained with $\lambda = \sqrt{n}$. Note that unlike previous results for WGANs \citep[e.g.][]{insights, without-domination}, our optimal rate of convergence does not depend on the (intrinsic or extrinsic) dimension of the dataset. This is because our rates quantify the speed at which the empirical risk of a distribution $\dist{g}$ reaches its population risk. In contrast, the usual rate of $n^{-1/d}$, where $d$ is the (intrinsic or extrinsic) dimension of the instance space $\xfancy$ quantifies the speed at which the population risk of the distribution $\dist{\hat{g}}$ minimizing the empirical problem $\min_{g\in \gfancy} \dfancy(\dist{*}_n, \dist{g})$ reaches the best possible performance $\inf_{g\in \gfancy} \dfancy(\dist{*}, \dist{g})$.

\section{Experiments}

\subsection{Preliminary Discussion}
Before presenting our experiments, we discuss some of the practical aspects of minimizing PAC-Bayesian bounds. First, we use probabilistic neural networks \citep{langford2001} with a Gaussian distribution on each parameter. %Minimizing a PAC-Bayesian bound requires to maintain a balance between the empirical risk and the KL divergence. 

\paragraph{Prior learning.} As illustrated by Equation~\eqref{eq-bound} the optimization of PAC-Bayes bounds requires a tradeoff between the empirical risk and the KL divergence $\kl{\rho}{\pi}$.  When using neural networks, controlling the KL divergence can be challenging, given the high dimensionality of the hypothesis class $\hfancy$ in that case. If the prior $\pi$ is independent from the data-generating distribution, then an optimal posterior $\rho$ is likely to be very far from $\pi$, leading to a KL divergence that is orders of magnitude larger than the empirical risk. To circumvent this issue, it is common in the PAC-Bayes literature \citep{perez-ortiz} to use a portion of the training set to learn the prior $\pi$. Given a training set of size $n$, the prior's mean is learned on $n_0 < n$ samples, the posterior $\rho$ is learned on all $n$ samples, and the bound in computed on the remaining $n-n_0$ samples. Both $\pi$ and $\rho$ have diagonal covariance matrices, and the prior's covariance matrix is chosen, whereas the posterior's is learned. Note that there are other strategies for choosing a PAC-Bayesian prior, such as fixing the mean vector to $\zeromatrix$ or random values from the standard normal distribution $\normal{\zeromatrix, \idmatrix}$. However, learning the prior usually leads to a more balanced optimization objective and tighter risk certificates. 

\paragraph{The impact of $\sigma_0$.} In our experiments the hyperparameter, $\sigma_0$ plays two roles. First, the prior $\pi$ is an isotropic Gaussian distribution with a covariance matrix $\sigma_0 \idmatrix$, and second, the initial value of the posterior's covariance matrix is also $\sigma_0 \idmatrix$. Note that the covariance matrix of the posterior $\rho$ is a learned diagonal matrix $\Sigma_\rho$, but we use $\sigma_0 \idmatrix$ as the initial value. Hence, $\sigma_0$ has a dual impact on the optimization. Since the KL divergence $\kl{\rho}{\pi}$ gets larger as the prior $\pi$ gets narrower, if $\sigma_0$ is too small, then the optimization may be too focused on the KL term, hence neglecting the empirical risk. However, because of the initial value of $\Sigma_\rho$, the variance of the posterior $\rho$ is likely to remain close to the variance of the prior, which helps control the KL divergence. On the other hand, if $\sigma_0$ is too large, then minimizing  $\kl{\rho}{\pi}$ may require the posterior $\rho$ to have a large variance as well, hence putting some weight on suboptimal generators and worsening the generative model's performance. This is illustrated in Figures \ref{fig-hist-ring} and \ref{fig-hist-grid}: when $\sigma_0 = 0.1$, the model's empirical and true risks are relatively large, compared to the other values of $\sigma_0$. Figures \ref{fig-fake-ring} and \ref{fig-fake-grid} in the appendix show samples generated from the different models. One can observe that for both synthetic datasets, when $\sigma_0 = 0.1$, the models do not learn the data-generated distribution well.

% \begin{itemize}
%     \item Minimizing PB bounds requires a trade-off between the empirical risk and the KL-divergence.
    
%     \item The prior learning trick we used is very common and allows one to control the KL divergence in order to avoid imbalance. 
    
%     \item The choice of $\sigma_0$ affects the optimization because ....
    
%     \item The additional cost of using our objective compared to the usual WGAN objective is very low since we optimize the Gibbs posterior. Meaning ...
    
%     \item Although the bounds for WGANs were stated for symmetric sets $\ffancy \subseteq \lipschitz{1}$, the results can be trivially extended to $\ffancy \subseteq \lipschitz{k}$, for any $k>0$ by scaling the critic loss. However, in order to compute the numerical values of the bounds, the Lipschitz constant of the critic network has to be known.
    
%     \item 
% \end{itemize}

\paragraph{Computational cost.} The additional cost of training using our objective is very low. Indeed, we optimize the Gibbs posterior during training, instead of  averaging over multiple generators $g \sampled \rho$. This means that at each iteration, we compute the empirical risk using samples from the training set $S$ and the distribution $\dist{g}$ given by a random generator $g \sampled \rho$. Moreover, since both the prior and the posterior are Gaussian distributions with diagonal covariance matrices, the KL divergence is easily computed \citep[see][Section~5.2]{perez-ortiz}.

\paragraph{Numerical computation of the bounds.} The numerical computation of our (non-desintegrated) bounds requires the empirical risk, the KL divergence, and an additional term dependent on the data-generating process (for instance, Equation~\eqref{eq-bound} requires the diameter of the instance space, while Equation~\eqref{eq-bound-lip-rand} requires the intrinsic dimension and the Lipschitz constant of $g^*$). For real-life datasets, both the intrinsic dimension and the smoothness of the data-generating process are unknown. Although there exists estimations of the former for some datasets \citep[e.g.][]{intrinsic-dim-images}, to the best of our knowledge, there are no estimations of the latter in the literature. Finally, note that although the bounds for WGANs assume the critic family $\ffancy \subseteq \lipschitz{1}$, in practice, once can still optimize the bounds and obtain risk certificates when the critic network's Lipschitz constant~$K$ is larger, since $f \in \lipschitz{1}$ if and only if $K f \in \lipschitz{K}$. Hence, in order to obtain valid risk certificates, one needs to scale the bounds accordingly, which requires the Lipschitz constant of the critic network to be known. This is not the case when using techniques such as the celebrated gradient penalty \citep{improved-wgan}.

\subsection{Synthetic datasets}

We perform experiments on two synthetic datasets: a mixture of 8~Gaussians arranged on a ring, and a mixture of 25~Gaussians arranged on a grid. These are standard synthetic datasets for GAN experiments, see, e.g, \citet{aligan, veegan, presgan}. In order to formally ensure the diameter of the instance space is finite, we truncate the data so that the first dataset is contained in a disc of radius $3.2$ and the second dataset in a square of side $8.2$, both centered at the origin. We optimized the right-hand side of Equation~\eqref{eq-bound} plus $\expect{g\sampled \rho}{ \newrisk{\dist{*}_n, \dist{g} }}$, estimating the latter expectation by randomly sampling 100 generators from $\rho$.
 In our chosen models, both the generator and critic are fully connected networks, and we use the Björk orthonormalization algorithm \citep{bjorck} to enforce Lipschitz continuity on the critic. We performed experiments using both ReLU and GroupSort activations \citep{sorting-out}, and we report the results using GroupSort as it leads to more stability.

The standard deviation of the prior $\pi$ is denoted $\sigma_0$ and we performed a sweep over the values $\sigma_0 \in \theset{10^{-7}, 10^{-6}, 10^{-5}, 0.0001, 0.001, 0.01, 0.1}$, and fix the hyperparameter $\lambda=\frac{n}{1024}$, where $n$ is the size of the training set. The standard deviation of the posterior is learned, and we use $\sigma_0$ as a starting point. Samples from the learned distributions are displayed in the appendix (Figures~\ref{fig-fake-ring} and~\ref{fig-fake-grid}).

\begin{figure}
    \centering
    \includegraphics[width=\linewidth, height=5.3cm]{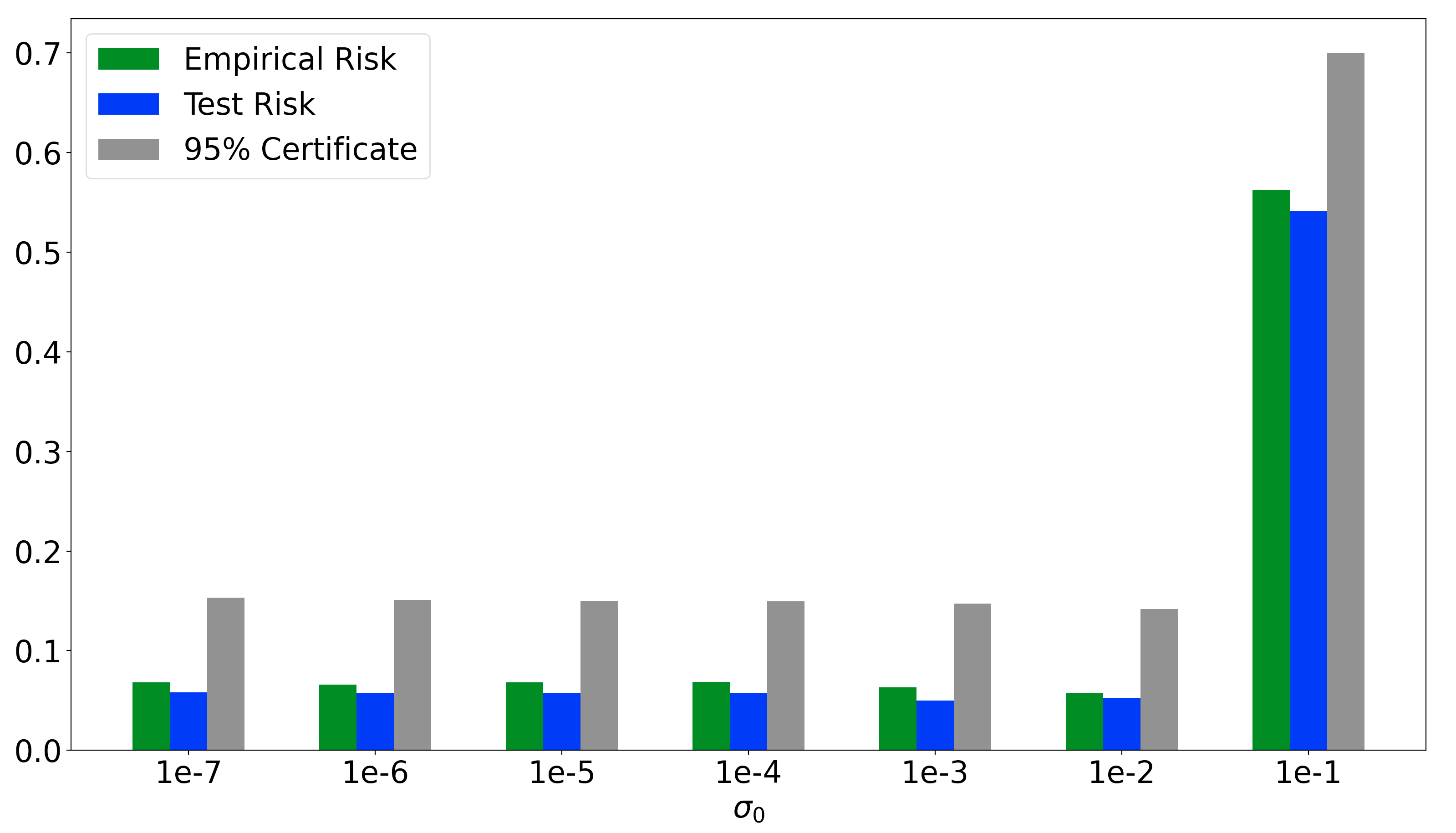}
    \caption{Negative critic losses and risk certificates of a model trained on a mixture of 8 Gaussian distributions arranged on a ring. The x-axis shows the value of the prior parameters' std $\sigma_0$. See Appendix (Fig.~\ref{fig-fake-ring}) for illustrations of the generated samples. }
    \label{fig-hist-ring}
\end{figure}

\begin{figure}
    \centering
    \includegraphics[width=\linewidth, height=5.3cm]{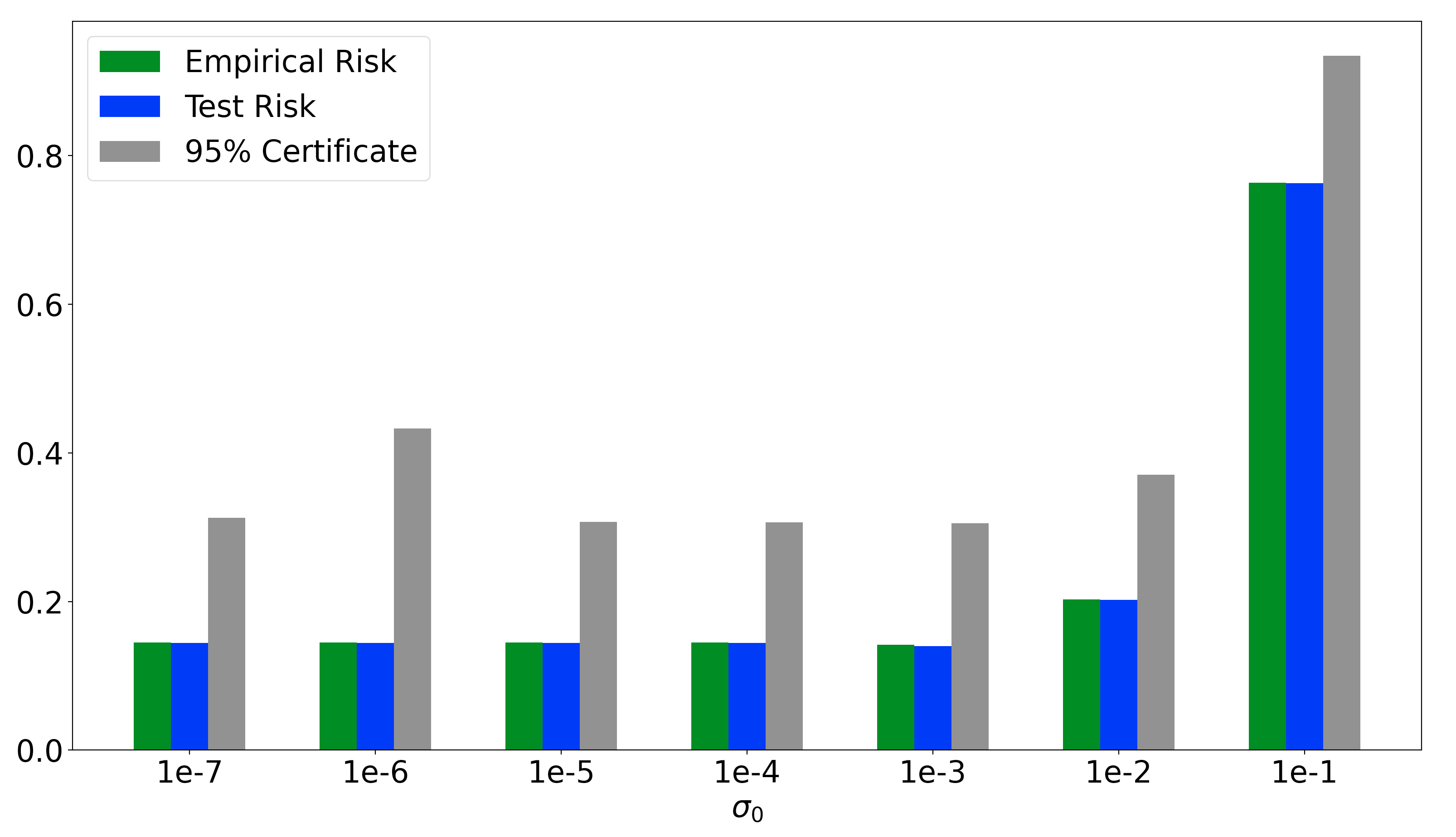}
    \caption{Negative critic losses and risk certificates of a model trained on a mixture of 25 Gaussian distributions arranged on a grid. The x-axis shows the value of the prior parameters' std $\sigma_0$. See Appendix (Fig.~\ref{fig-fake-grid}) for illustrations of the generated samples. }
    \label{fig-hist-grid}
\end{figure}

Figures \ref{fig-hist-ring} and \ref{fig-hist-grid} show the risks (negative critic losses) on the training and the test sets, as well as the risk certificate given by Equation~\eqref{eq-bound}, for the different values of the hyperparameter $\sigma_0$. The expectations with respect to $g\sampled \rho$ are approximated by averaging over $100$ generators independently sampled from~$\rho$.

We observe that the learned generator has similar empirical and test risks. This is a known asset of learning by optimizing a PAC-Bayesian bound, as it prevents overfitting the training samples.
%\citep{...}. 
We even notice that some model instances have an empirical risk slightly larger than their test risk, a phenomenon rarely observed when training a discriminative (prediction) model. In our generative setting, this indicates that
the critic's ability to distinguish the real samples from the fake ones is consistent, whether the real samples are from the training set or the test set. The computed risk certificates lie in the same order of magnitude than the test loss, which qualifies them as \textit{non-vacuous}.

%(a term that has been coined in the PAC-Bayes literature by \citet{...}). 
% Of note, the bound value provides an accurate model selection criteria for the parameter $\sigma_0$, \emph{i.e.}, the model that has the lowest bound comes with the best test risk. This is true even if the model with $\sigma_0 =10^{-3}$ in the \emph{grid} experiment of Figure~\ref{fig-hist-grid} comes with a bound value that is less tight than for other parameter values (relatively to their test risk). This indicates that the models have to reach a higher complexity (measured by the term $\kl{\rho}{\pi}$ of~\eqref{eq-bound}) to drive down the empirical risk, and illustrates the trade-off captured by Theorem~\ref{thm-pb-bound}.

\subsection{Experiments on MNIST}
We performed preliminary experiments on the MNIST dataset \citep{mnist} using the standard DCGAN architecture \citep{dcgan}, which requires the images to be re-sized to 64 x 64 pixels. Here, we used gradient penalty \citep{improved-wgan} to enforce Lipschitz continuity on the critic. Similar to the experiments on synthetic datasets, we used different values of $\sigma_0$ and computed the FID scores on 2000 random samples from each model. See Section~\ref{sec-app-mnist} for more details.

%\vspace{2cm}
% The goal is to study the correlation between the numerical value of the bounds and the quality of the samples. We perform a grid sweep over the values of $\sigma_0 \in \theset{10^{-7}, 10^{-6}, 10^{-5}, 0.0001, 0.001, 0.01}$ and $\sigma_\rho \in \theset{10^{-7}, 10^{-6}, 10^{-5}, 0.0001, 0.001, 0.01}$, where $\sigma_0$ is the standard deviation of the prior's parameters and $\sigma_\rho$ is the starting point of the standard deviation of the posterior's parameters $\sigma_1, \dots, \sigma_P$. Note that we do learn $\sigma_1, \dots, \sigma_P$, but we use $\sigma_i = \sigma_\rho$ as a starting point for the optimization.

% Since the diameter ... and Theorem ... requires the Lipschitz constant $K$ of the data-generating process, which is unknown, we compute $\beta(model) = ...$ for the different models and perform a comparison ... 

%  \details{Explain architecture details and choices and say we're using Thm 3.4. Explain that the term with $K$ is not needed, since it doesn't depend on individual models, and we don't know the Lipschitz constant of the critic anyway, since we're using GP.}

\section{Conclusion and Future Works}

Recent years have seen a growing interest in PAC-Bayesian theory, as a framework for deriving statistical guarantees for a variety of machine learning models \citep{primer}. Despite the long list of topics for which PAC-Bayesian bounds have been developed, generative models were missing from this list. In this work, we developed PAC-Bayesian bounds for adversarial generative models. We showed that these bounds can be numerically computed and provide non-vacuous risk certificates for synthetic datasets. 

In future works, we will explore risk certificates on real-life datasets. Unlike synthetic datasets for which we can have all the information such as the intrinsic and extrinsic dimensions, real-life datasets come with the challenge that some information is unknown. Computing the bounds of Theorem~\ref{thm-pb-bound} would require the use of the diameter of the instance space, which is clearly irrelevant to the structure of the dataset. On the other hand, the bounds of Theorem~\ref{thm-pb-bound-lipschitz} require some information about the smoothness of the data generating process. In future works, we will explore empirical estimations of that quantity.

\section*{Acknowledgements}

This research is supported by the Canada CIFAR AI Chair
Program, and the NSERC Discovery grant RGPIN-2020-07223.
F.~Clerc is funded by IVADO through the DEEL project and by a grant from NSERC.


\bibliography{ref}
\bibliographystyle{icml2023}

%%%%%%%%%%%%%%%%%%%%%%%%%%%%%%%%%%%%%%%%%%%%%%%%%%%%%%%%%%%%%%%%%%%%%%%%%%%%%%%
%%%%%%%%%%%%%%%%%%%%%%%%%%%%%%%%%%%%%%%%%%%%%%%%%%%%%%%%%%%%%%%%%%%%%%%%%%%%%%%
% APPENDIX
%%%%%%%%%%%%%%%%%%%%%%%%%%%%%%%%%%%%%%%%%%%%%%%%%%%%%%%%%%%%%%%%%%%%%%%%%%%%%%%
%%%%%%%%%%%%%%%%%%%%%%%%%%%%%%%%%%%%%%%%%%%%%%%%%%%%%%%%%%%%%%%%%%%%%%%%%%%%%%%
\newpage
\appendix
\onecolumn

\section{Proofs}

\subsection{Preliminaries}

We start this section with the following definition.
\begin{definition}[Bounded differences] \label{def-bounddif}
 A function $f : \xfancy^n \rightarrow \real$ is said to have the \emph{bounded differences property} if for some non-negative constants $c_1, \dots, c_n$, we have for any $1 \leq i \leq n$,
\[
\sup_{x_1, \dots, x_n, x_i' \in \xfancy} \absolu{f(x_1, \dots, x_n) - f(x_1, \dots, x_{i-1}, x_i', x_{i+1}, \dots, x_n)} \leq c_i.
\]
In other words, if we change the $i^{th}$ argument of $f$ while keeping all the others fixed, the value of the function cannot change by more than $c_i$. 
\end{definition}

\noindent
The following lemma is used to prove a special case of McDiarmid's inequality \citep{mcdiarmid}. 
%Although this result is not explicitely stated in McDiarmid's article \citep{mcdiarmid}, it is often attributed to the latter.
\begin{lemma}\label{lem-mcdiarmid}
Let $f : \xfancy^n \rightarrow \real$ be a function that has the bounded differences property with constants $c_i, 1 \leq i \leq n$. Then, denoting $Z = f(\listen{\xvector})$, we have that
\begin{equation}%\label{}
 \expect{}{\exponbig{\lambda \left(  \expect{}{Z} - Z \right)}}  \leq \exponbig{\lambda^2\nu / 8},
\end{equation}
where $\nu = \somme{i=1}{n}{c_i^2}$.

\end{lemma}

\noindent
Below, we include a summary of the proof by \citet{mcdiarmid-note} (with minor modifications) for completeness.

\begin{proof}
The proof relies on the clever use of the following functions: for each $1 \leq k \leq n$, we define a function $g_k : \xfancy^k \rightarrow \real$ by
\begin{align*}
    g_k(\xvector_1, \dots, \xvector_k) & = \expect{\xvector_{k}, \dots, \xvector_n}{f(\listen{\xvector})} - \expect{\xvector_{k+1}, \dots, \xvector_n}{f(\listen{\xvector})}, \text{ when } k< n,\\
    g_n(\xvector_1, \dots, \xvector_n) &= \expect{\xvector_{n}}{f(\listen{\xvector})} - f(\listen{\xvector}).
\end{align*}
For every $k$, the function $g_k$ satisfies the following results:
\[ \expect{\xvector_k}{g_k(\xvector_1, \dots, \xvector_k)} = 0 \qquad \text{and} \qquad
0 \leq b_k - a_k \leq c_k,\]
where we have denoted $a_k = \inf_{\xvector_k} g_k(\xvector_1, \dots, \xvector_k)$ and $b_k = \sup_{\xvector_k} g_k(\xvector_1, \dots, \xvector_k)$. These results allow us to conclude using Hoeffding's lemma that for every $k$
\[ \int_\xfancy \expon{\lambda g_k(\xvector_1, \dots, \xvector_k)}~ d\dist{*}(\xvector_k) \leq \expon{\lambda^2 c_k^2 / 8}. \]
Finally, we use the fact that
\[ \expect{}{Z} - Z = \somme{k=1}{n}{g_k(\xvector_1, \dots, \xvector_k)} \]
to rewrite $\expect{}{\expon{\lambda \left( \expect{}{Z} - Z   \right)}    }$ using Fubini's thorem. We get the desired result by induction using previously-shown inequality.
\end{proof}

\subsection{Proof of Theorem \ref{thm-pb-bound}}
\label{sec-thm-pb-bound}

\begin{lemma}\label{lem-w1-bounddif}
Let $P, Q$ be probability measures on $\xfancy$ and $P_n, Q_n$ be the empirical distributions corresponding to the iid samples $\listen{\xvector} \sampled P$ and $\listen{\yvector} \sampled Q$ respectively, meaning
\[
P_n (\xvector) = \unsur{n} \somme{i=1}{n}{\delta_{\xvector_i}(\xvector)} \andspace
Q_n (\xvector) = \unsur{n} \somme{i=1}{n}{\delta_{\yvector_i}(\xvector)}
\]
for any $\xvector \in \xfancy$. 

Let $\ffancy \subseteq \lipschitz{1}$ be a symmetric subset of $\lipschitz{1}$. Recall the definition of the IPM defined by $\ffancy$:
\[
\ipm{\ffancy}(P, Q) = \sup_{f\in \ffancy} \left\{ \integral{}{ }{f}{dP} - \integral{ }{ }{f}{dQ}   \right\}.
\] 
Then the empirical IPM $\ipm{\ffancy}(P_n, Q_n)$, seen as a function $\xfancy^{2n} \rightarrow \real$, has the bounded differences property with $c_i = \frac{\Delta}{n}$ and $\Delta=\rm\diam{\xfancy}$, for all $1 \leq i \leq 2n$. 
\end{lemma}

\begin{proof}
 We show, without loss of generality, that $c_n = \frac{\Delta}{n}$. We have
\begin{equation*}
\setlength{\jot}{10pt}
\begin{split}
& \ipm{\ffancy}(\listen{\xvector}, \listen{\yvector}) - \ipm{\ffancy}(\listen{\xvector}', \listen{\yvector})  \\
    & = \unsur{n}  \left\{ \supremum{f\in \ffancy}{\left[  \somme{i=1}{n}{f(\xvector_i)}  -  \somme{i=1}{n}{f(\yvector_i)}  \right]} -  \supremum{f \in \ffancy}{\left[  \somme{i=1}{n-1}{f(\xvector_i)} + f(\xvector_n') - \somme{i=1}{n}{f(\yvector_i)}  \right]}  \right\}        \\
    & \leq \unsur{n}  \left\{ \supremum{f\in \ffancy}{\left[  \somme{i=1}{n}{f(\xvector_i)}  -  \somme{i=1}{n}{f(\yvector_i)}  -  \somme{i=1}{n-1}{f(\xvector_i)} - f(\xvector_n') + \somme{i=1}{n}{f(\yvector_i)}  \right] } \right\}               \\
    & = \unsur{n} \sup_{f\in \ffancy} \left[  f(\xvector_n) -  f(\xvector_n')   \right]  \\
    & \leq  \frac{\Delta}{n}.       \\
\end{split}
\end{equation*}
The first inequality (second to third lines) follows from a property of the supremum and the last inequality follows from $\ffancy \subseteq \lipschitz{1}$ and $\diam{\xfancy}{=}\Delta$.
\end{proof}
When $\ffancy = \lipschitz{1}$, then Lemma \ref{lem-w1-bounddif} states that the Wasserstein distance between empirical measures has the bounded differences property, which follows from a result by \citet{weed_bach}. 

\medskip

\noindent
Combining Lemmas \ref{lem-mcdiarmid} and \ref{lem-w1-bounddif} yields the following result.

\begin{proposition}\label{prop-expon-moment}
Let $P$ and $Q$ be two probability measures on $\xfancy$ and $P_n, Q_n$ be their empirical counterparts  corresponding to $S_P$ and $S_Q$ respectively. Then
\[
\expect{ }{\exponbig{\lambda \left(  \expect{ }{\ipm{\ffancy}(P_n, Q_n)} - \ipm{\ffancy}(P_n, Q_n)      \right)}    } \leq \exponbig{\frac{\lambda^2 \Delta^2}{4n}},
\]
where both expectations are taken over $(S_P, S_Q) \sampled P^{\otimes n} \times Q^{\otimes n}$.
\end{proposition}

\noindent
We are now ready to prove Theorem \ref{thm-pb-bound}.

\begin{proof}[Proof of Theorem \ref{thm-pb-bound}] \mbox{}
\begin{enumerate}[label=(\roman*), leftmargin=*]
\item 
For a given generator $g \in \gfancy$, Proposition \ref{prop-expon-moment} implies
\[
\expectseul{S, S_g}{\exponbig{\lambda \left(  \expect{S, S_g}{\ipm{\ffancy}(\dist{g}_n, P^*_n)}   - \ipm{\ffancy}(\dist{g}_n, P^*_n)   \right)}} \leq \exponbig{\frac{\lambda^2 \Delta^2}{4n}},
\]
where we write $\expectseul{S, S_g}{}$ instead of $\expectseul{(S, S_g) \sampled {\dist{*}}\otimesn \times {\dist{g}}\otimesn}{}$ in order to simplify the notation.
Taking the average with respect to the prior $\pi \in \mprob{\gfancy}$ and using Fubini's theorem, we get
\begin{equation}\label{eq-expect-leq}
 \expectseul{S }{
\expectseul{g \sampled \pi}{
\expect{S_g}{
\exponbig{\lambda \left(  \expect{S, S_g}{\ipm{\ffancy}(\dist{g}_n, P^*_n)}   - \ipm{\ffancy}(\dist{g}_n, P^*_n)   \right)}     
}
}     
} \leq \exponbig{\frac{\lambda^2 \Delta^2}{4n}}.
\end{equation}
Now, defining
\[
Y \equdef \expectseul{g \sampled \pi}{
\expect{S_g}{
\exponbig{\lambda \left(  \expect{S, S_g}{\ipm{\ffancy}(\dist{g}_n, P^*_n)}   - \ipm{\ffancy}(\dist{g}_n, P^*_n)   \right)}     
}
},
\]
we have that $Y$ is a positive random variable and Markov's inequality implies
\[
\prob{}{Y \geq \unsur{\delta} \expect{}{Y}} \leq \delta
\]
for any real number $\delta \in (0, 1)$. Taking complementary event, we get that with probability at least $1-\delta$ over the random draw of $S \sampled {\dist{*}}\otimesn$, 
\[
Y \leq \unsur{\delta} \expect{}{Y} \leq \unsur{\delta}\exponbig{\frac{\lambda^2 \Delta^2}{4n}},
\]
where the last inequality follows from \eqref{eq-expect-leq}. So we've just shown that with probability at least $1-\delta$ over the random draw of the training set $S \sampled {\dist{*}}\otimesn$, 
\[
\expectseul{g \sampled \pi}{
\expect{S_g}{
\exponbig{\lambda \left(  \expect{S, S_g}{\ipm{\ffancy}(\dist{g}_n, P^*_n)}   - \ipm{\ffancy}(\dist{g}_n, P^*_n)   \right)}     
}
}     \leq 
\unsur{\delta}\exponbig{\frac{\lambda^2 \Delta^2}{4n}}. 
\]
Now, assume $\rho \in \mprob{\gfancy}$ is such that $\pi \abscont \rho$ and $\rho \abscont \pi$. We can change the expectation with respect to $\pi$ into an expectation with respect to $\rho$ using the Radon-Nikodym derivative $\frac{d\pi}{d\rho}$ to obtain
\[
\expect{g \sampled \rho}{ \frac{d\pi}{d\rho}
\expect{S_g}{
\exponbig{\lambda \left(  \expect{S, S_g}{\ipm{\ffancy}(\dist{g}_n, P^*_n)}   - \ipm{\ffancy}(\dist{g}_n, P^*_n)   \right)}     
}
}     \leq 
\unsur{\delta}\exponbig{\frac{\lambda^2 \Delta^2}{4n}}. 
\]
Taking the logarithm on both sides and using Jensen's inequality yields
\[
\expect{g \sampled \rho}{ 
\expect{S_g}{ 
\log \left(\frac{d\pi}{d\rho}\right) + 
\lambda \left(  \expect{S, S_g}{\ipm{\ffancy}(\dist{g}_n, P^*_n)}   - \ipm{\ffancy}(\dist{g}_n, P^*_n)   \right)
}
}     \leq 
\log \unsur{\delta} +\frac{\lambda^2 \Delta^2}{4n},
\]
which is equivalent to
\[
- \expect{g \sampled \rho}{ \log \left(\frac{d\rho}{d\pi}\right)  } + \expect{g \sampled \rho}{ 
\expect{S_g}{  
\lambda \left(  \expect{S, S_g}{\ipm{\ffancy}(\dist{g}_n, P^*_n)}   - \ipm{\ffancy}(\dist{g}_n, P^*_n)   \right)
}
}     \leq 
\log \unsur{\delta} +\frac{\lambda^2 \Delta^2}{4n},
\]
since $\rho \abscont \pi$ and  $\frac{d\pi}{d\rho} = \left(  \frac{d\rho}{d\pi}\right)^{-1} $. This last inequality can be re-written as follows:
\[
\lambda \left(  
\expectseul{g \sampled \rho}{ \expect{S_g}{  
\expect{S }{\ipm{\ffancy}(\dist{g}_n, P^*_n)}   - \ipm{\ffancy}(\dist{g}_n, P^*_n)  
}
} \right)  \leq 
\kl{\rho}{\pi} + \log \unsur{\delta} +\frac{\lambda^2 \Delta^2}{4n},
\]
or, using the linearity of the expectation and the definition $\newrisk{\dist{*}_n, \dist{g}} = \expect{S_g}{ \ipm{\ffancy}(\dist{*}_n, \dist{g}_n) }$,
\[
\lambda \left(  
\expectseul{g \sampled \rho}{ 
\expect{S }{\newrisk{\dist{*}_n, \dist{g}}} 
}  -
\expectseul{g \sampled \rho}{  \newrisk{\dist{g}_n, P^*_n}
}  \right)  \leq 
\kl{\rho}{\pi} + \log \unsur{\delta} +\frac{\lambda^2 \Delta^2}{4n}. 
\]
The proof above uses the ideas of \citet{pac-bayes-linear} and \citet{unleashed}. We provided details for completeness and clarity.

\item Denote
\begin{equation*} \label{eq-xi}
\xi = \log \expectseul{S}{\expectseul{g \sampled \pi}{\expect{S_g}{\expon{    \lambda \left(   \expectseul{S}{ \expect{S_g}{\ipm{\ffancy} (\dist{*}_n, \dist{g}_n) } }  - { \ipm{\ffancy} (\dist{*}_n, \dist{g}_n) } \right)     }}}}.
\end{equation*}
First, using Fubini's theorem and Proposition \ref{prop-expon-moment} we have
\begin{equation*}
\setlength{\jot}{10pt}
\begin{split}
\xi &=  \log \expectseul{g \sampled \pi}{\expectseul{S}{\expect{S_g}{\expon{    \lambda \left(   \expectseul{S}{\expect{S_g}{\ipm{\ffancy} (\dist{*}_n, \dist{g}_n) } }  - { \ipm{\ffancy} (\dist{*}_n, \dist{g}_n) } \right)     }}}}          \\
   & \leq \log \expect{g \sampled \pi}{\expon{\frac{\lambda ^2 \Delta^2}{4n}}}          \\
   & = \frac{\lambda ^2 \Delta^2}{4n} .         
\end{split}
\end{equation*}
Then, using the convexity of the exponential and Jensen's inequality, we obtain
\begin{equation*}
\setlength{\jot}{10pt}
\begin{split}
\xi &=  \log \expectseul{S}{\expectseul{g \sampled \pi}{\expect{S_g}{\expon{    \lambda \left(   \expectseul{S}{ \expect{S_g}{\ipm{\ffancy} (\dist{*}_n, \dist{g}_n) } }  - { \ipm{\ffancy} (\dist{*}_n, \dist{g}_n) } \right)     }}}}         \\
   &  \geq   \log \expectseul{S}{\expectseul{g \sampled \pi}{\expon{    \lambda \expectseul{S_g}{ \left(   \expectseul{S}{ \expect{S_g}{\ipm{\ffancy} (\dist{*}_n, \dist{g}_n) } }  - { \ipm{\ffancy} (\dist{*}_n, \dist{g}_n) } \right)     }}}}            \\
   & =  \log \expectseul{S}{\expectseul{g \sampled \pi}{\expon{    \lambda \left(   \expectseul{S}{ \expect{S_g}{\ipm{\ffancy} (\dist{*}_n, \dist{g}_n) } }  - { \expectseul{S_g}{ \ipm{\ffancy} (\dist{*}_n, \dist{g}_n) }} \right)     }}}.
\end{split}
\end{equation*}
The combination of these two inequalities yields
\begin{equation}\label{eq-xi-combine}
\log \expectseul{S}{\expectseul{g \sampled \pi}{\expon{    \lambda \left(   \expectseul{S}{ \expect{S_g}{\ipm{\ffancy} (\dist{*}_n, \dist{g}_n) } }  - { \expectseul{S_g}{ \ipm{\ffancy} (\dist{*}_n, \dist{g}_n) }} \right)     }}} \leq \frac{\lambda ^2 \Delta^2}{4n} .
\end{equation}
Now, a result by \citet{beyond-usual} states that for any measurable function $f$, the following holds with probability at least $1-\delta$ over the random draw of $S \sampled {\dist{*}} \otimesn$ and $g \sampled \rho$:
\begin{equation*}\label{}
  f(S, g) \leq \log \frac{d\rho}{d\pi}(g) + \log \unsur{\delta} + \log \expectseul{S}{\expect{g \sampled \pi}{\expon{f(S, g)}}}.
\end{equation*}
Taking 
\[
f(S, g) = \lambda \left(   \expect{S, S_g}{\ipm{\ffancy} (\dist{*}_n, \dist{g}_n)  }  - \expect{S_g}{ \ipm{\ffancy} (\dist{*}_n, \dist{g}_n) } \right),
\]
we get
\begin{equation*}\label{eqn-exp-diff-expected}
 \begin{split}
 \lambda \left(   \expect{S, S_g}{\ipm{\ffancy} (\dist{*}_n, \dist{g}_n)  }  - \expect{S_g}{ \ipm{\ffancy} (\dist{*}_n, \dist{g}_n) } \right) 
 \leq 
\log \frac{d\rho}{d\pi}(g) + \log \unsur{\delta} +  \\  \log \expectseul{S}{\expectseul{g \sampled \pi}{\expon{    \lambda \left(   \expect{S, S_g}{\ipm{\ffancy} (\dist{*}_n, \dist{g}_n)  }  - \expect{S_g}{ \ipm{\ffancy} (\dist{*}_n, \dist{g}_n) } \right)   }  }  }.
\end{split}
\end{equation*}
Combining this result with \eqref{eq-xi-combine}, we obtain
\[
 \lambda \left(   \expect{S, S_g}{\ipm{\ffancy} (\dist{*}_n, \dist{g}_n)  }  - \expect{S_g}{ \ipm{\ffancy} (\dist{*}_n, \dist{g}_n) } \right) 
 \leq 
\log \frac{d\rho}{d\pi}(g) + \log \unsur{\delta} + \frac{\lambda ^2 \Delta^2}{4n} .
\]

\end{enumerate}
\end{proof}

\paragraph{Remark.} As stated in the main paper, increasing the number of fake samples $S_g$ from $n$ to $m$ worsens the bounds. This is because in that case, the constants $c_i = \frac{\Delta}{n}$ of Lemma~\ref{lem-w1-bounddif} become $c_i = \max(\frac{\Delta}{n}, \frac{\Delta}{m})$, leading to a worse bound.

Next, we prove Corollary \ref{cor-bound}. 
\begin{proof}[Proof of Corollary \ref{cor-bound}]
Denote $S = \theset{\listen{\xvector}}$ and $S_g = \theset{\listen{\yvector}}$ the iid datasets corresponding to the empirical distributions $\dist{*}_n$ and $\dist{g}_n$ respectively. The properties of the supremum imply
\[
\supremum{f \in \ffancy} \left\{   \expect{S, S_g}{  \integral{}{}{f }{d\dist{g}_n}  -  \integral{}{}{f }{d\dist{*}_n}    }   \right\} \leq
\expect{S, S_g}{\supremum{f \in \ffancy}{ \left\{ \integral{}{}{f }{d\dist{g}_n}  -  \integral{}{}{f }{d\dist{*}_n}   \right\} }  } .
\]
Moreover, since $S$ and $S_g$ are iid datasets, we have
\[
\expect{S}{\unsur{n}\somme{i=1}{n}{f(\xvector_i)}} = \expect{\xvector \sampled P^*}{f(\xvector)} \andspace
\expect{S_g}{\unsur{n}\somme{i=1}{n}{f(\yvector_i)}} = \expect{\yvector \sampled \dist{g}}{f(\yvector)}.
\]
Therefore,
\[
\ipm{\ffancy}(\dist{*}, \dist{g})
= \supremum{f\in \ffancy}{\expect{S, S_g}{\unsur{n} \somme{i=1}{n}{f(\xvector_i)} -\unsur{n} \somme{i=1}{n}{f(\yvector_i)} }   }
\leq \expectseul{S, S_g}{\supremum{f\in \ffancy}{ \left\{ \unsur{n} \somme{i=1}{n}{f(\xvector_i)} -\unsur{n} \somme{i=1}{n}{f(\yvector_i)} \right\} }   } = \expectseul{S, S_g}{\ipm{\ffancy}(\dist{*}_n, \dist{g}}).
\]
Combining this inequality with Theorems \ref{thm-pb-bound}-\ref{thm-pb-bound-rand} and \ref{thm-pb-bound}-\ref{thm-pb-bound-derand} yields the desired results.
\end{proof}

\subsection{Proof of Theorem \ref{thm-pb-bound-lipschitz}}
\label{sec-thm-pb-bound-lipschitz}
The proof of Theorem \ref{thm-pb-bound-lipschitz} is similar to the proof of Theorem \ref{thm-pb-bound}. The only difference is that instead of Lemma \ref{lem-w1-bounddif}, we use the following result.

\begin{lemma} \label{lem-w1-bounddif-lipschitz}
Let $\zfancy = [0, 1]^\dimz$ and $P_\zfancy$ be a probability measure on $\zfancy$. Let  $P, Q$ be probability measures on $\xfancy$ such that $P=\pushf{g_1}{P_\zfancy}$ and $Q=\pushf{g_2}{P_\zfancy}$ with $g_1, g_2 \in \lipschitz{K}$, $K \geq 1$. Let $P_n, Q_n$ be the empirical distributions corresponding to the iid samples $\listen{\xvector} \sampled P$ and $\listen{\yvector} \sampled Q$ respectively. Then the function $\wprime: \xfancy^{2n} \rightarrow \real$, defined as
\[
\wprime (\listen{\xvector}, \listen{\yvector}) = \wprime(P_n, Q_n),
\]
has the bounded differences property with $c_i = \frac{K\sqrt{\dimz}}{n}$, for all $1 \leq i \leq 2n$.
\end{lemma}

\begin{proof}
First, let $\listen{w}, w_n', \listen{z} \sampled P_\zfancy$ such that for all $1 \leq i \leq n$,
\begin{equation} \label{eq-xi-wi-lip}
\xvector_i = g_1(w_i), \,\, \xvector_n' = g_1(w_n') \andspace \yvector_i = g_2(\zvector_i).
\end{equation}
 We have
\begin{equation*}
\setlength{\jot}{10pt}
\begin{split}
& \wprime(\listen{\xvector}, \listen{\yvector}) - \wprime(\listen{\xvector}', \listen{\yvector})  \\
    & = \unsur{n}  \left\{ \supremum{f\in \ffancy}{\left[  \somme{i=1}{n}{f(\xvector_i)}  -  \somme{i=1}{n}{f(\yvector_i)}  \right]} -  \supremum{f \in \ffancy}{\left[  \somme{i=1}{n-1}{f(\xvector_i)} + f(\xvector_n') - \somme{i=1}{n}{f(\yvector_i)}  \right]}  \right\}        \\
    & \leq \unsur{n}  \left\{ \supremum{f\in \ffancy}{\left[  \somme{i=1}{n}{f(\xvector_i)}  -  \somme{i=1}{n}{f(\yvector_i)}  -  \somme{i=1}{n-1}{f(\xvector_i)} - f(\xvector_n') + \somme{i=1}{n}{f(\yvector_i)}  \right] } \right\}               \\
    & = \unsur{n} \sup_{f\in \ffancy} \left[  f(\xvector_n) -  f(\xvector_n')   \right]  \\
    & \leq   \frac{K \sqrt{\dimz}}{n} .       \\
\end{split}
\end{equation*}
In order to prove the last inequality, we just need to show that for any $f \in \ffancy$, $f(\xvector_n) - f(\xvector_n') \leq K \sqrt{\dimz}$.
Let $f \in \ffancy$. Using~\eqref{eq-xi-wi-lip} and the assumptions $\ffancy \subseteq \lipschitz{1}$, $g_1 \in \lipschitz{K}$ and $\zfancy=[0, 1]^{\dimz}$, which implies $\diam{\zfancy} = \sqrt{\dimz}$, we have
\[
f(\xvector_n) - f(\xvector_n') = f(g_1(w_n)) - f(g_1(w_n'))  \leq K \sqrt{\dimz}.
\]
\end{proof}

The following result is similar to Corollary \ref{cor-bound} and bounds the distance between the full distributions.

\begin{corollary}\label{cor-bound-lipschitz}
With the definitions and assumptions of Theorem~\ref{thm-pb-bound-lipschitz}, the following properties hold for any probability measure $\rho$ such that $\rho \abscont \pi$ and $\pi \abscont \rho$.
\begin{enumerate}[label=(\roman*)]

\item \label{cor-rand2} With probability at least $1-\delta$ over the random draw of $S$:
\begin{equation}
\expectseul{g\sampled \rho}{\ipm{\ffancy}(\dist{*}, \dist{g}) }  \leq \expect{g\sampled \rho}{ \newrisk{\dist{*}_n, \dist{g} } } +  
\unsur{\lambda }     \left[ \kl{\rho}{\pi}   +\log \unsur{\delta}   \right] + \frac{\lambda K^2\dimz}{4n}.
\end{equation}

\item \label{cor-derand2} With probability at least $1-\delta$ over the random draw of $S$ and $g \sampled \rho$:
\begin{equation}
\ipm{\ffancy}(\dist{*}, \dist{g})     \leq \newrisk{\dist{*}_n, \dist{g}} + 
\unsur{\lambda }     \left[ \log\frac{d\rho}{d\pi}(g) +\log \unsur{\delta}   \right] + \frac{\lambda K^2\dimz}{4n}.
\end{equation}
\end{enumerate}
\end{corollary}

\subsection{Proof of Theorem \ref{thm-pb-bound-tv}}
\label{sec-thm-pb-bound-tv}
The proof of Theorem \ref{thm-pb-bound-tv} requires the following result.

\begin{lemma} \label{lem-w1-bounddif-tv}
Let  $P, Q$ be probability measures on $\xfancy$ and $P_n, Q_n$ be the empirical distributions corresponding to the iid samples $\listen{\xvector} \sampled P$ and $\listen{\yvector} \sampled Q$ respectively. Then the empirical total variation distance 
has the bounded differences property with $c_i = \frac{2}{n}$, for all $1 \leq i \leq 2n$.
\end{lemma}

\begin{proof}
We have
\begin{equation*}
\setlength{\jot}{10pt}
\begin{split}
& \tvprime(\listen{\xvector}, \listen{\yvector}) - \tvprime(\listen{\xvector}', \listen{\yvector})  \\
    & = \unsur{n}  \left\{ \supremum{f\in \ffancy}{\left[  \somme{i=1}{n}{f(\xvector_i)}  -  \somme{i=1}{n}{f(\yvector_i)}  \right]} -  \supremum{f \in \ffancy}{\left[  \somme{i=1}{n-1}{f(\xvector_i)} + f(\xvector_n') - \somme{i=1}{n}{f(\yvector_i)}  \right]}  \right\}        \\
    & \leq \unsur{n}  \left\{ \supremum{f\in \ffancy}{\left[  \somme{i=1}{n}{f(\xvector_i)}  -  \somme{i=1}{n}{f(\yvector_i)}  -  \somme{i=1}{n-1}{f(\xvector_i)} - f(\xvector_n') + \somme{i=1}{n}{f(\yvector_i)}  \right] } \right\}               \\
    & = \unsur{n} \sup_{f\in \ffancy} \left[  f(\xvector_n) -  f(\xvector_n')   \right]  \\
    & \leq   \frac{2}{n} .       \\
\end{split}
\end{equation*}
The last inequality follows from $-1 \leq f \leq 1$, for any $f \in \ffancy$.
\end{proof}

The following result is similar to Corollaries \ref{cor-bound} and \ref{cor-bound-lipschitz}. It  bounds on the distance between the full distributions.

\begin{corollary}\label{cor-bound-tv}
With the definitions and assumptions of Theorem~\ref{thm-pb-bound-tv}, the following properties hold for any probability measure $\rho$ such that $\rho \abscont \pi$ and $\pi \abscont \rho$.
\begin{enumerate}[label=(\roman*)]

\item \label{cor-rand3} With probability at least $1-\delta$ over the random draw of $S$:
\begin{equation}
\expectseul{g\sampled \rho}{\ipm{\ffancy}(\dist{*}, \dist{g}) }  \leq \expect{g\sampled \rho}{ \tvprime(\dist{*}_n, \dist{g} ) } +  
\unsur{\lambda }     \left[ \kl{\rho}{\pi}   +\log \unsur{\delta}   \right] + \frac{4\lambda}{n}.
\end{equation}

\item \label{cor-derand3} With probability at least $1-\delta$ over the random draw of $S$ and $g \sampled \rho$:
\begin{equation}
\ipm{\ffancy}(\dist{*}, \dist{g})     \leq \tvprime(\dist{*}_n, \dist{g} )  + 
\unsur{\lambda }     \left[ \log\frac{d\rho}{d\pi}(g) +\log \unsur{\delta}   \right] + \frac{4\lambda}{n}.
\end{equation}
\end{enumerate}
\end{corollary}

%%%%%%%%%%%%%%%%%%%%%%%%%%%%%%%%%%%%%%%%%%%%%%%
%%%%%%%%%%%%%%%%%%%%%%%%%%%%%%%%%%%%%%%%%%%%%%%

\section{Samples from the experiments} 

\subsection{Synthetic datasets}
\begin{figure}
    \centering
    \includegraphics[width=.6\linewidth, height=5cm]{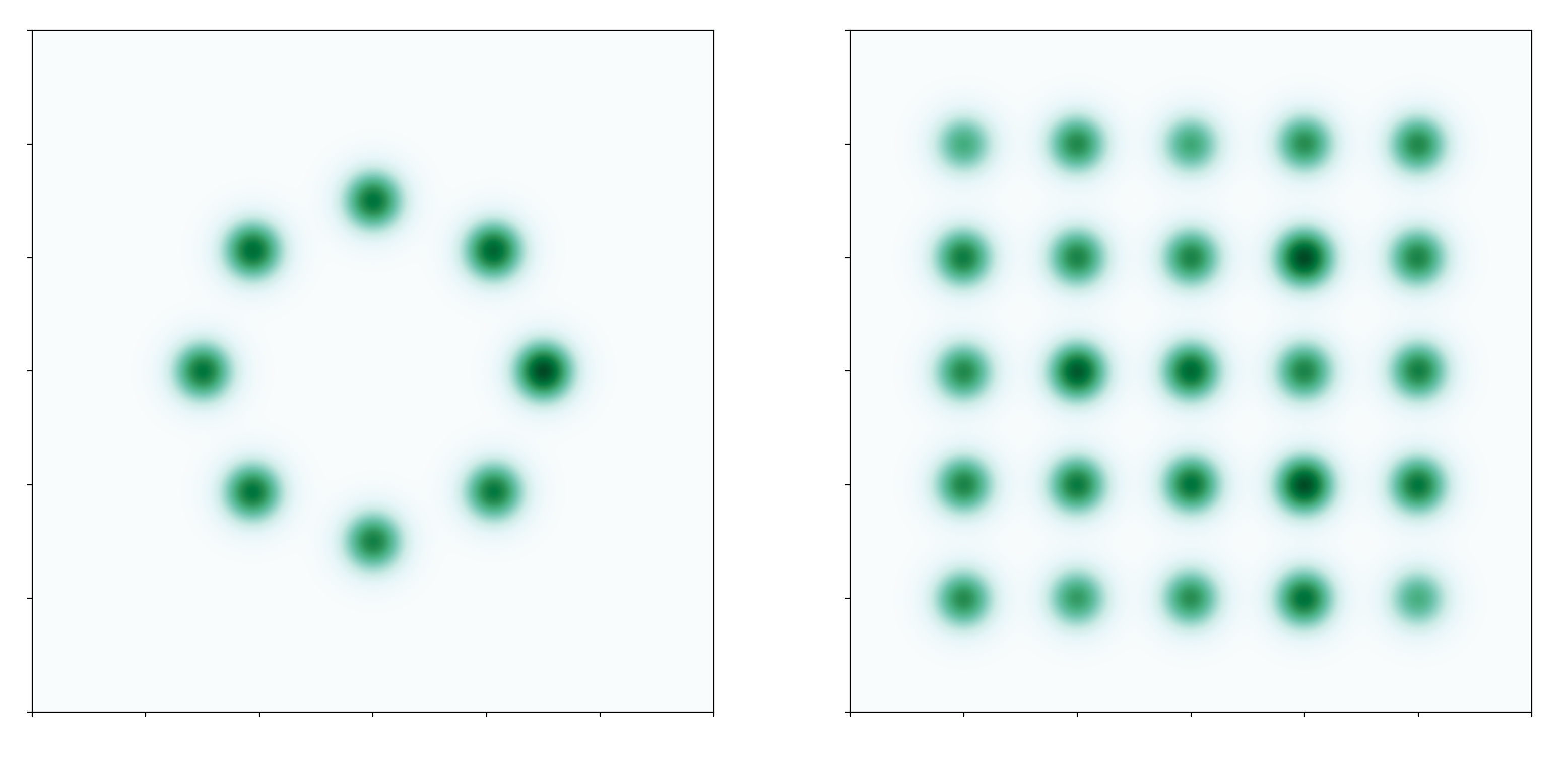}
    \caption{Real samples from the respective datasets. The image on the left represents samples from the Gaussian mixture with 8 components arranged on a ring, and the image on the right shows samples from the Gaussian mixture with 25 components arranged on a grid.}
    \label{fig-real-samples}
\end{figure}

\begin{figure}
    \centering
    \includegraphics[width=.85\linewidth, height=2.7cm]{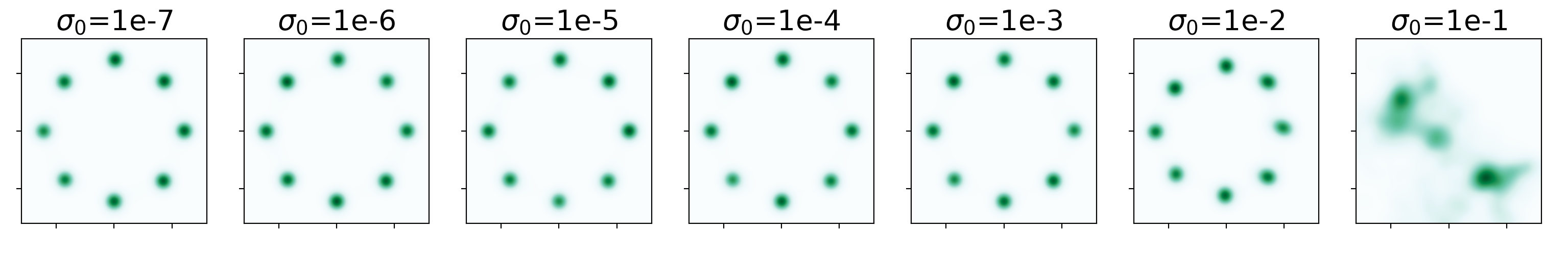}
    \caption{Samples from the models trained on the Gaussian ring}
    \label{fig-fake-ring}
\end{figure}

\begin{figure}
    \centering
    \includegraphics[width=.8\linewidth, height=2.7cm]{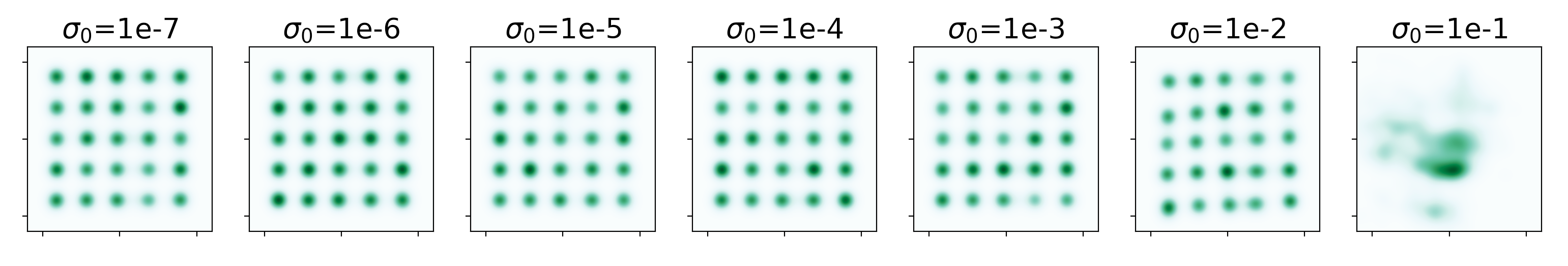}
    \caption{Samples from the models trained on the Gaussian grid}
    \label{fig-fake-grid}
\end{figure}

We used two datasets: a Gaussian mixture with eight components arranged on a ring, and a Gaussian mixture with nine components on a grid. Figure \ref{fig-real-samples} shows real samples from the actual training sets, and Figures \ref{fig-fake-ring} and \ref{fig-fake-grid} show samples from the  trained models.

\subsection{MNIST dataset}\label{sec-app-mnist}
\begin{figure}[h!]
\centering
\begin{minipage}{.47\linewidth}
  \centering
  \includegraphics[width=\linewidth]{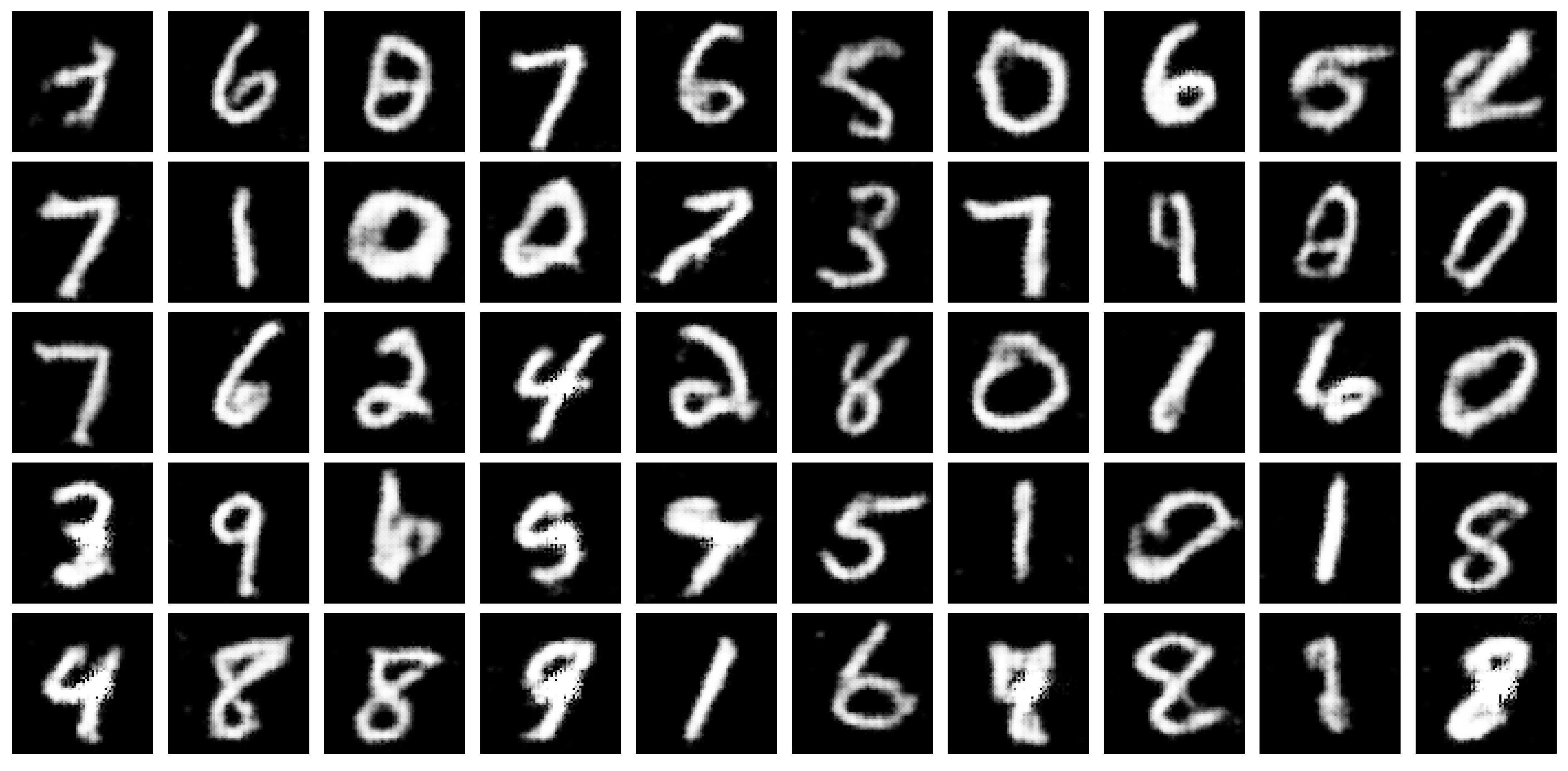}
  \caption{$\sigma_0 = 10^{-2}$}
  \label{fig-mnist-2}
\end{minipage}% 
\hfill
\begin{minipage}{.47\linewidth}
  \centering
  \includegraphics[width=\linewidth]{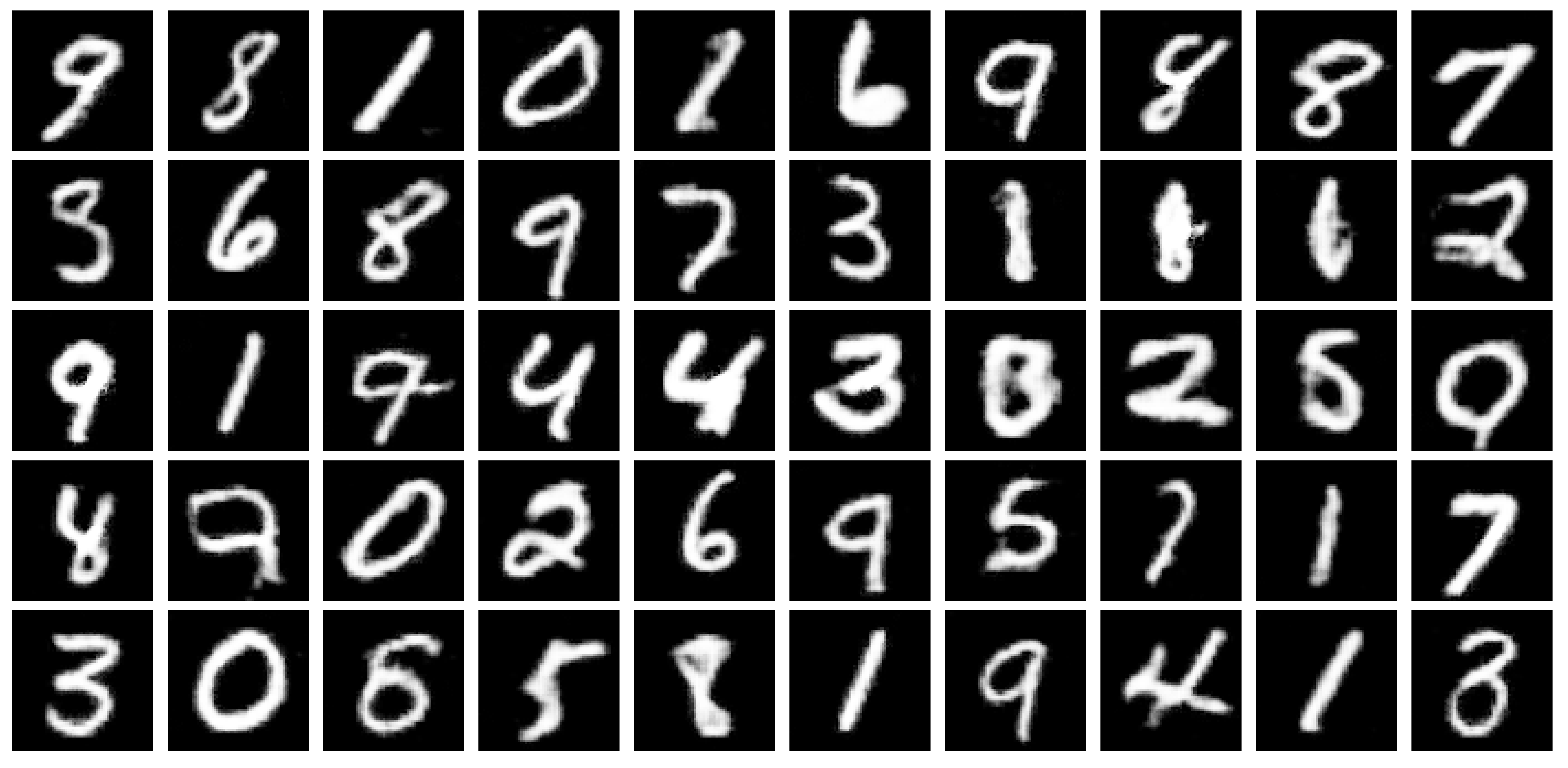}
  \caption{$\sigma_0 = 10^{-3}$}
  \label{fig-mnist-3}
\end{minipage}
\end{figure}

\begin{figure}[h!]
\centering
\begin{minipage}{.47\linewidth}
  \centering
  \includegraphics[width=\linewidth]{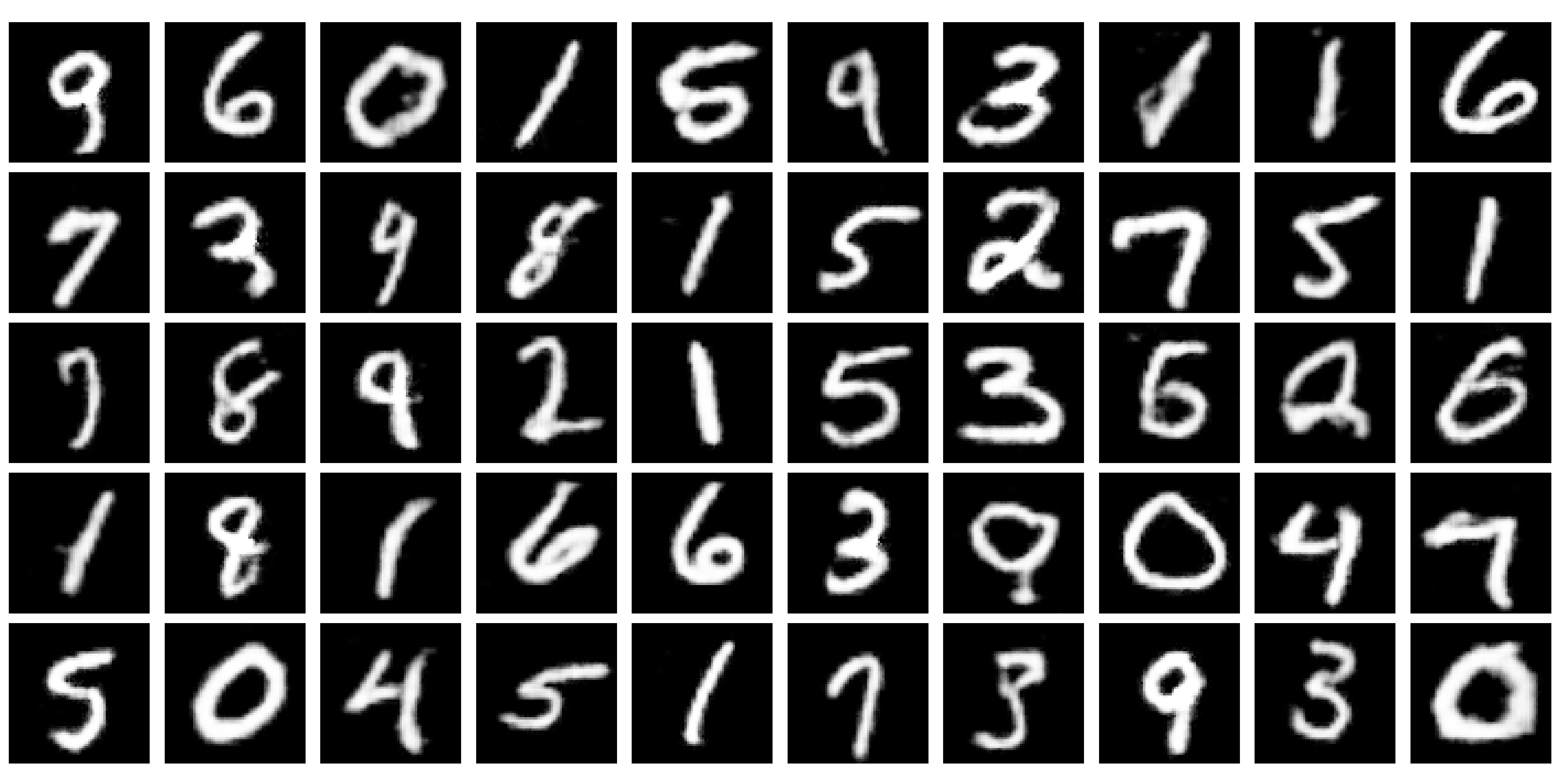}
  \caption{$\sigma_0 = 10^{-4}$}
  \label{fig-mnist-4}
\end{minipage}% 
\hfill
\begin{minipage}{.47\linewidth}
  \centering
  \includegraphics[width=\linewidth]{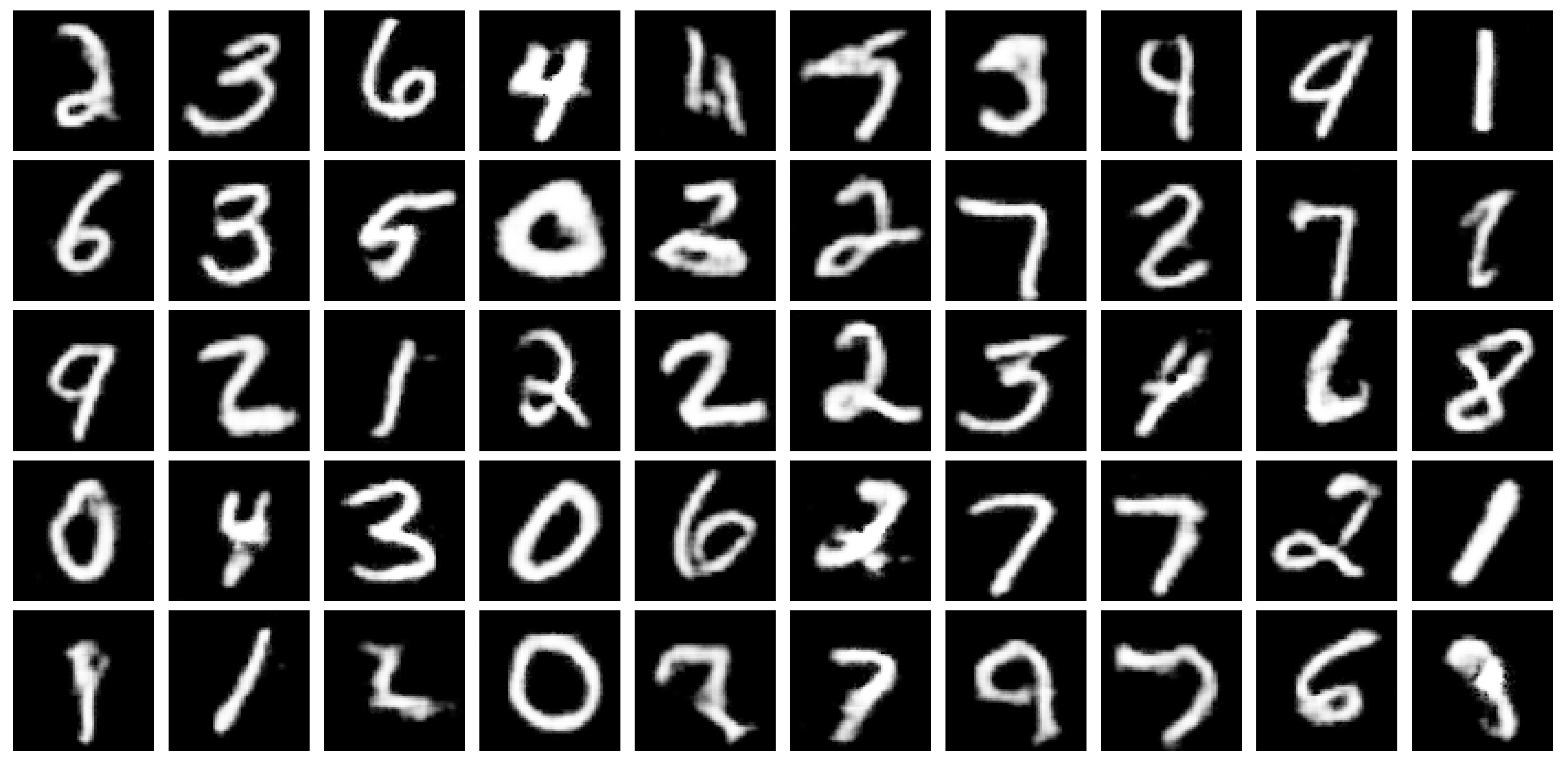}
  \caption{$\sigma_0 = 10^{-5}$}
  \label{fig-mnist-5}
\end{minipage}
\end{figure}

\begin{figure}[h!]
\centering
\begin{minipage}{.47\linewidth}
  \centering
  \includegraphics[width=\linewidth]{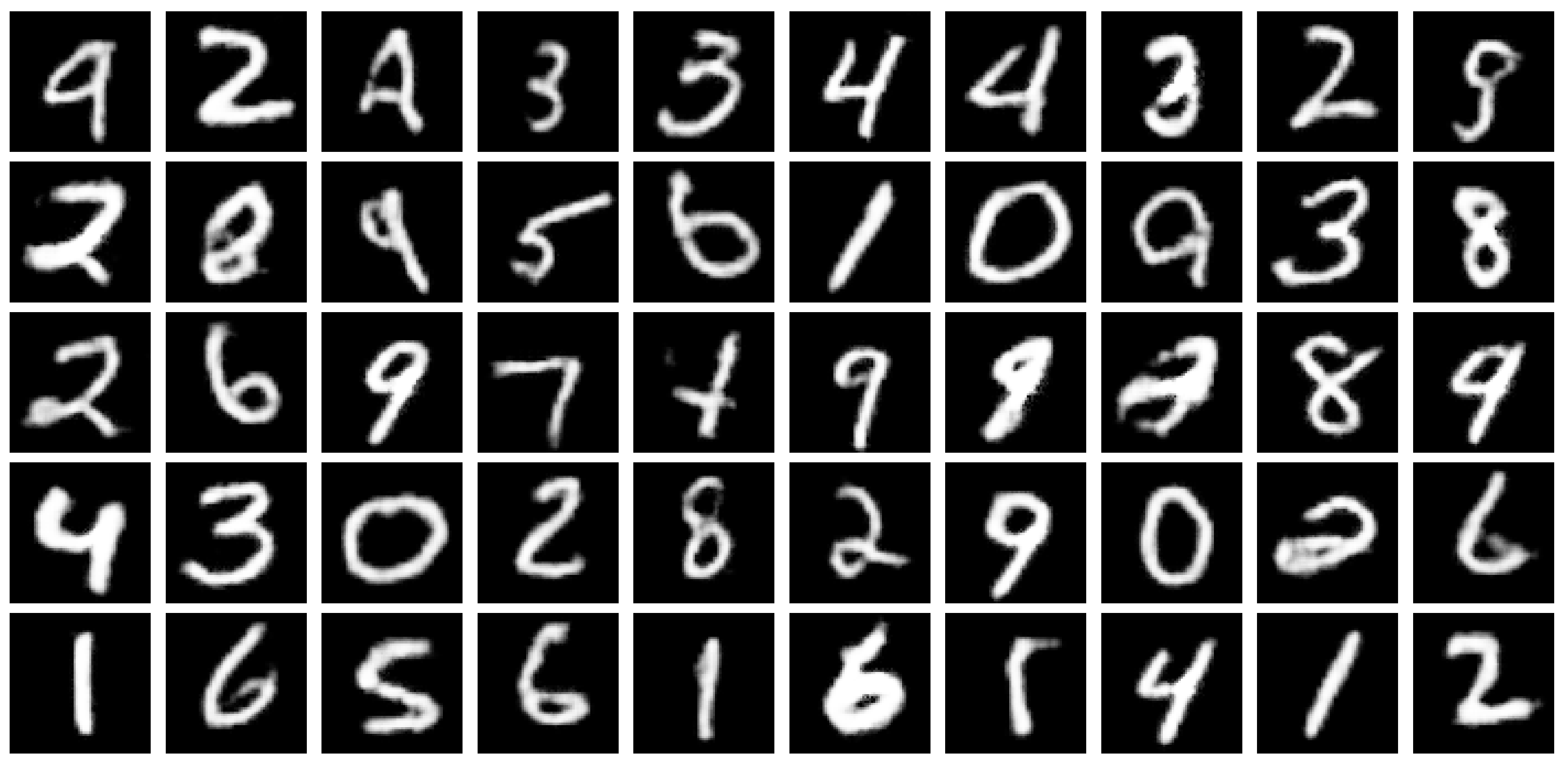}
  \caption{$\sigma_0 = 10^{-6}$}
  \label{fig-mnist-6}
\end{minipage}% 
\hfill
\begin{minipage}{.47\linewidth}
  \centering
  \includegraphics[width=\linewidth]{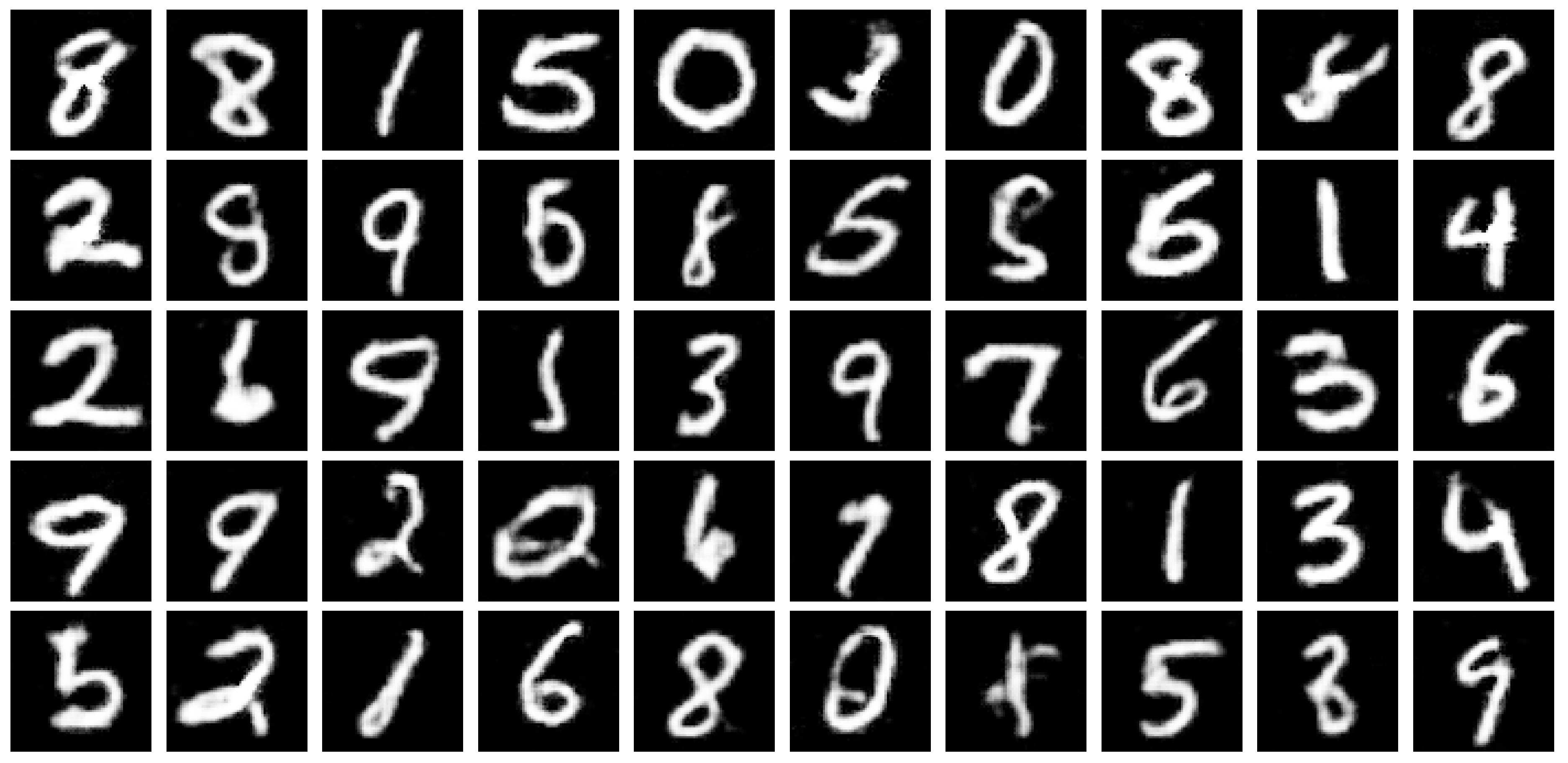}
  \caption{$\sigma_0 = 10^{-7}$}
  \label{fig-mnist-7}
\end{minipage}
\end{figure}

In our experiments with the MNIST dataset \citep{mnist}, we used the standard DCGAN architecture \citep{dcgan} for the generator and the critic. We experimented with different values for the hyperparameter $\sigma_0$ to train the probabilistic models. We computed the FID scores \citep{fid-score} for the different models using $2000$ random samples and the off-the-shelf implementation provided in the Pytorch-ignite library \citep{pytorch-ignite}, with a inception network \citep{inception-net} pre-trained on Imagenet. Since the Inception network requires 3-channel images, we transformed the original MNIST images by copying the single channel twice. The scores obtained for different models are displayed in Table~\ref{tab-fid} and random (not cherry-picked) samples are displayed on Figures \ref{fig-mnist-2} to \ref{fig-mnist-7}.

\begin{table}[t]\label{tab-fid}
\caption{FID scores from the various models trained on MNIST}
\label{sample-table}
\vskip 0.15in
\begin{center}
\begin{small}
\begin{sc}
\begin{tabular}{cc}
\toprule
$\sigma_0$ & Score \\
\midrule
$10^{-7}$    & 113.16  \\
$10^{-6}$    & 106.59  \\
$10^{-5}$    & 111.15  \\
$10^{-4}$    & 107.54  \\
$10^{-3}$    & 112.51  \\
$10^{-2}$    & 189.30  \\
\bottomrule
\end{tabular}
\end{sc}
\end{small}
\end{center}
\vskip -0.1in
\end{table}

%%%%%%%%%%%%%%%%%%%%%%%%%%%%%%%%%%%%%%%%%%%%%%%%%%%%%%%%%%%%%%%%%%%%%%%%%%%%%%%
%%%%%%%%%%%%%%%%%%%%%%%%%%%%%%%%%%%%%%%%%%%%%%%%%%%%%%%%%%%%%%%%%%%%%%%%%%%%%%%

\end{document}